\documentclass[runningheads]{llncs}

\usepackage{tikz}
\usepackage{pgfplots}
\usepackage{graphicx}
\usepackage{amsmath,amssymb} 
\usepackage{color}
\usepackage{booktabs}
\usepackage{usr}
\usepackage{tabularx}
\usepackage{cite}
\usepackage{hyperref}
\pgfplotsset{compat=1.13}
\begin{document}

\newcommand{\todo}[1]{{\textbf{\color{red}#1}}}
\newcommand{\bmat}[1]{\begin{bmatrix}#1\end{bmatrix}}
\newcommand{\pmat}[1]{\begin{pmatrix}#1\end{pmatrix}}

\newcommand{\TS}[1]{{\textbf{\color{blue}Torsten: #1}}}

\newcommand{\PAR}[1]{\vskip4pt \noindent{\bf #1~}}

\pagestyle{headings}
\mainmatter
\def\ECCVSubNumber{2550}  

\title{Infrastructure-based Multi-Camera Calibration using Radial Projections} 




\titlerunning{Infrastructure-based Multi-Camera Calibration using Radial Projections}
%
\author{Yukai Lin\inst{1} \and
Viktor Larsson\inst{1} \and
Marcel Geppert\inst{1} \and
Zuzana Kukelova\inst{2} \and
Marc Pollefeys\inst{1,3} \and
Torsten Sattler\inst{4}}
\authorrunning{Y. Lin et al.}
%
\institute{Department of Computer Science, ETH Z\"urich, \and
VRG, Faculty of Electrical Engineering, Czech Technical University in Prague \and
Microsoft Mixed Reality \& AI Z\"urich Lab \and
Department of Electrical Engineering, Chalmers University of Technology
}

\maketitle

\begin{abstract}
Multi-camera systems are an important sensor platform for intelligent systems such as self-driving cars. Pattern-based calibration techniques can be used to calibrate the intrinsics of the cameras individually. However, extrinsic calibration of systems with little to no visual overlap between the cameras is a challenge. Given the camera intrinsics, infrastucture-based calibration techniques are able to estimate the extrinsics using 3D maps pre-built via SLAM or Structure-from-Motion. In this paper, we propose to fully calibrate a multi-camera system from scratch using an infrastructure-based approach. Assuming that the distortion is mainly radial, we introduce a two-stage approach. We first estimate the camera-rig extrinsics up to a single unknown translation component per camera. Next, we solve for both the intrinsic parameters and the missing translation components. Extensive experiments on multiple indoor and outdoor scenes with multiple multi-camera systems show that our calibration method achieves high accuracy and robustness. In particular, our approach is more robust than the naive approach of first estimating intrinsic parameters and pose per camera before refining the extrinsic parameters of the system. 
The implementation is available at \href{https://github.com/youkely/InfrasCal}{\textit{https://github.com/youkely/InfrasCal}}.
\end{abstract}

\section{Introduction}
\begin{figure}[t]
\centering
\subfigure[Pentagonal camera rig]{
\label{fig:10cam-image}
\includegraphics[trim=80 40 85 0,clip,width=0.42\textwidth]{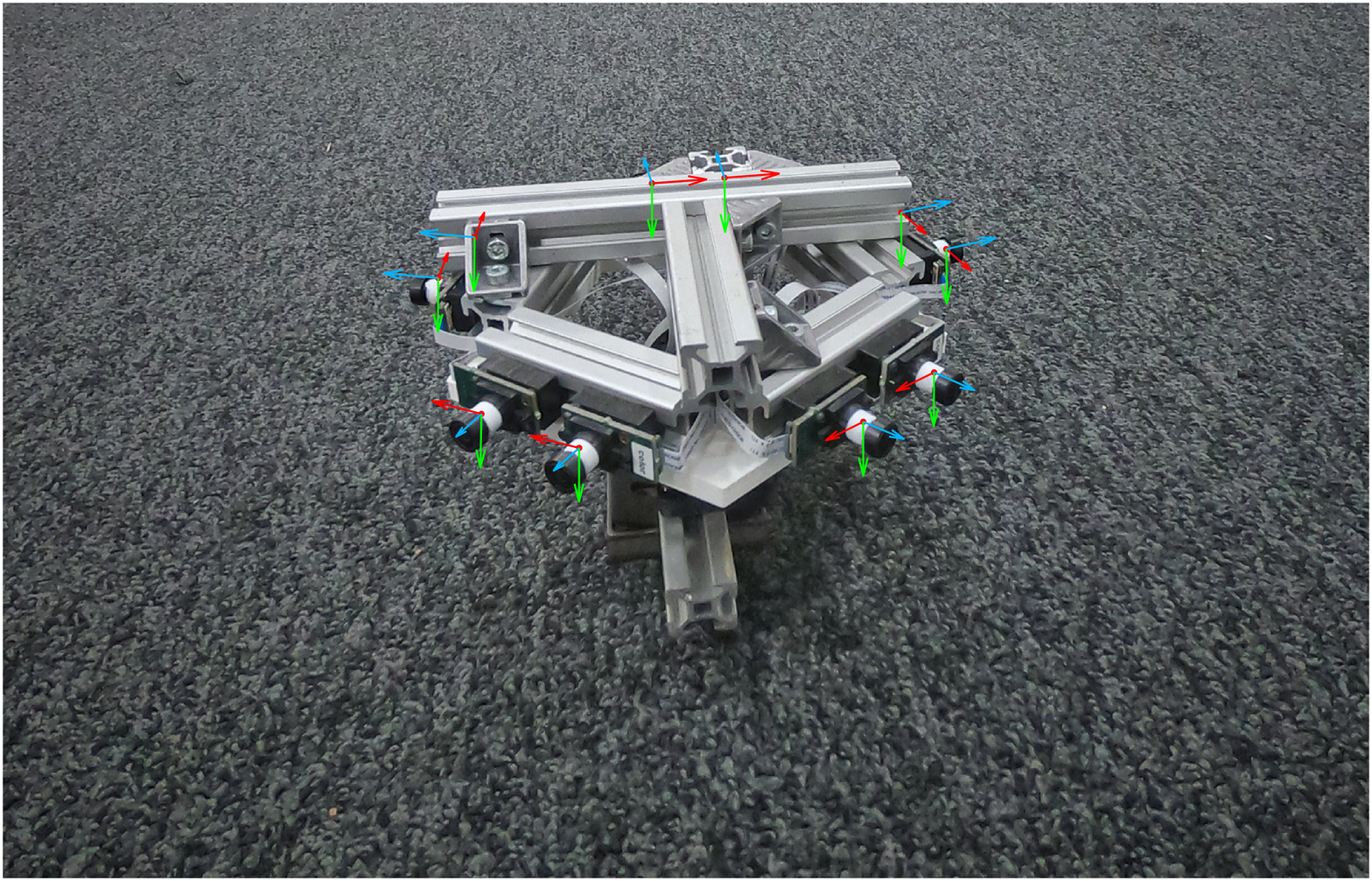}}
\subfigure[GoPro helmet]{
\label{fig:gopro-image}
\includegraphics[trim=80 30 70 0,clip,width=0.42\textwidth]{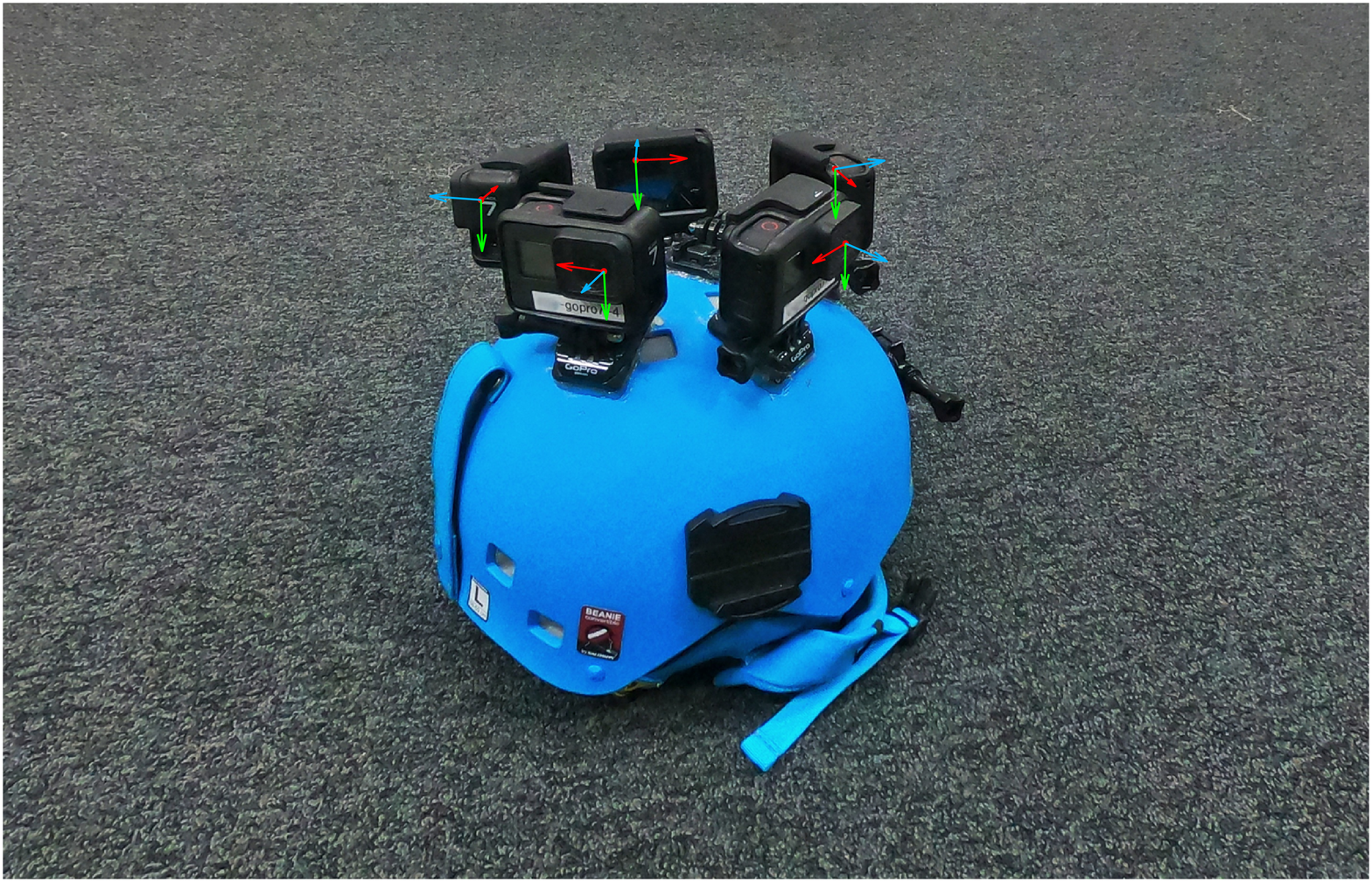}}
\caption{\textbf{Multi-camera rigs}. Our estimated rig calibrations are overlayed in the figure. (a) Pentagonal camera rig with five stereo pairs. (b) Ski helmet with five GoPro Hero7 Black attached covering 360$^\circ$ panoramic view. }
\label{fig:rig-image}
\end{figure}
Being able to perceive the surrounding environment is a crucial ability for any type of autonomous intelligent system, including self-driving cars~\cite{schwesinger2016automated,heng2019project} and robots~\cite{strisciuglio2018trimbot2020}. 
Multi-camera systems (see \eg Figure~\ref{fig:rig-image}) are popular sensors for this task: they are cheap to build and maintain, consume little energy, and provide high-resolution data under a wide range of conditions. 
Enabling 360$^\circ$ perception around a vehicle~\cite{schwesinger2016automated} using such systems, makes visual localization~\cite{arth2009wide,geppert2019efficient} and SLAM~\cite{liu2018towards} more robust. 

Multi-camera systems need to be calibrated before use. This includes calibrating the intrinsic parameters of each camera, \ie focal length, principal point and distortion parameters, as well as the extrinsic parameters between cameras, \ie, the relative poses between them. %
An accurate and efficient calibration is often crucial for safe and robust performance.
A standard approach to this problem, implemented in calibration toolboxes such as Kalibr~\cite{maye2013self}, is to use a calibration pattern to record data which covers the full \gls{fov} of the cameras. 
Although this method is powerful in achieving high accuracy, it is computationally expensive and recording a calibration dataset with adequate motion/view coverage is cumbersome, especially for wide-FoV cameras.
Moreover, it is incapable of calibrating the camera-rigs with little or even no overlapping fields of view, which is often the case for applications in autonomous vehicles.

Another approach to handle such scenarios are sequence- and infrastructure-based calibration~\cite{heng2015leveraging,heng2015self}.  
In both cases, the methods require prior knowledge of the intrinsics before the extrinsics calibration, which still requires a per camera pre-calibration step using calibration patterns.

In this paper, we introduce an infrastructure-based calibration that calibrates both intrinsics and extrinsics in a single pipeline. 
Our method uses a pre-build map of sparse feature points as a substitute for the calibration patterns.
The map is easily built by a Structure-from-Motion pipeline, \eg COLMAP~\cite{schonberger2016structure}. 
We calibrate the camera-rigs in a two-stage process. 
In the first stage, the camera poses are estimated under a radial camera assumption, where the extrinsics are recovered up to an unknown translation along the principal axis. 
In the second stage, the intrinsics and the remaining translation parameters are jointly estimated in a robust way. 
We demonstrate the accuracy and robustness through extensive experiments in indoor and outdoor datasets with different multi-camera systems.

The \textbf{main contributions} of this paper are: 
(\textbf{1})    We propose an infrastructure-based calibration method for performing multi-camera rig intrinsics and extrinsics calibration in an user-friendly way as we remove the need for pre-calibration for each camera or tedious recording for calibration pattern data.
(\textbf{2})    In contrast to current methods, we show that it is possible to first (partially) estimate the camera rig's extrinsic parameters before estimating the internal calibration for each camera. 
(\textbf{3})    Our proposed method is experimentally shown to give high-quality camera calibrations in a variety of environments and hardware setups.

\section{Related Work}
\label{sec:radialpose}

\PAR{Pattern-Based Calibration.}
A pattern-based calibration method estimates camera parameters using special calibration patterns such as AprilTags~\cite{olson2011apriltag} or checkerboards~\cite{maye2013self,sturm1999plane,zhang2004extrinsic}. 
The patterns are precisely designed so that they can be accurately estimated via camera systems. 
We note that the pattern-based calibration of multi-camera systems usually requires the camera pairs to have overlapping \gls{fov}s, since the pattern must be visible in multiple images to constrain the rig's extrinsic parameters.
Some works~\cite{kumar2008simple,li2013multiple,robinson2017robust} calibrate the cameras without assuming the overlapping \gls{fov}.
Kumar et al.~\cite{kumar2008simple} show that the use of an additional mirror can help to create overlap between cameras.
Li et al.~\cite{li2013multiple} only require the neighboring cameras to partially observe the calibration patterns at the same time but the observed parts do not necessarily need to overlap.
Robinson et al.~\cite{robinson2017robust} calibrate the extrinsic parameters for non-overlapping cameras by temporarily adding an additional camera during calibration with an overlapping \gls{fov} with both cameras. 
We note that the use of a calibration pattern board always introduces a certain viewing constraint or extra effort to calibrate the cameras with non-overlapping \gls{fov}.
Furthermore, the calibration of wide \gls{fov} cameras is especially cumbersome.
The pattern needs to be close to the camera to cover any significant part of the image but if it is too close, it leads to problems where the pattern is out of focus. Thus, to get accurate calibration results, it is typically necessary to capture a large number of images. 

\PAR{Infrastructure-Based Calibration.} 
Rather than using calibration patterns, infrastructure-based calibration uses natural scene features to estimate camera parameters.
Carrera et al.~\cite{carrera2011slam} propose a feature-based extrinsic calibration method through a SLAM-based feature matching among the maps for each camera.
Heng et al.~\cite{heng2015leveraging} simplify that approach to rely on a prior high-accuracy map, removing the need for inter-camera feature correspondences and loop closures.
Their pipeline first infers camera poses via the P3P method for calibrated cameras, and
subsequently, an initial estimate of the camera-rig transformations and rig poses. A final non-linear refinement step optimizes the camera-rig transformations, rig poses and optionally intrinsics.

Our method is most similar to the work of Heng et al.~\cite{heng2015leveraging} in that we use a pre-built sparse feature map for calibration.
However, their method relies on a known intrinsics input which still requires calibration patterns for intrinsic calibration. Our method does not require a prior intrinsics knowledge and performs complete calibration, both intrinsic and extrinsic, using the sparse map.

Compared to checkerboard-style calibration objects, infrastructure-based methods are able to get significantly more constraints per-image since there are typically more feature points observed which acts as a virtual large calibration pattern. 
In practice, infrastructure-based calibration provides a much wider application range than pattern-based calibration.

\PAR{Camera Pose Estimation with Unknown Intrinsic Parameters.}
Given a sparse set of 2D-3D correspondences between an image and a 3D point cloud (a map),  it is possible to recover the camera pose. If the cameras' internal calibration is known, i.e.~the mapping from image pixels to viewing rays, the absolute pose estimation problem becomes minimal with three correspondences. This problem is usually referred to as the Perspective-Three-Points (P3P) problem~\cite{haralick1994review}. In settings where the intrinsic parameters are unknown, the estimation problem becomes more difficult and more correspondences are necessary. Most modern cameras can be modeled as having square pixels (i.e.~zero skew and unit aspect ratio). Due to this, most work on camera pose estimation with unknown/partially known calibration has focused on the case of unknown focal length. The minimal problem for this case was originally solved by Bujnak et al.~\cite{bujnak2008general}. Since then, there have been several papers improving on the original solver \cite{penate2013exhaustive,zheng2014general,wu2015p3,kukelova2016efficient,larsson2017making}.
The case of unknown focal length and principal point was considered by Triggs \cite{triggs1999camera} and later Larsson et al.~\cite{larsson2018camera}. When all of the intrinsic parameters are unknown, the Direct-Linear-Transform (DLT)~\cite{hartley2003multiple} can be applied. 
Camera pose estimation with unknown radial distortion was first considered by Josephson and Byr\"od~\cite{josephson2009pose}. There have been multiple works improving on this paper in different aspects; faster runtime \cite{kukelova2013real,larsson2017making}, support for other distortion models \cite{larsson2019revisiting} and even non-parametric distortion models \cite{camposeco2015non}.

\PAR{Radial Alignment Constraint and the 1D Radial Camera Model.} 
Focal length and radial distortion only scales the images points radially outwards from the principal point (assuming this is the center of distortion). This observation was used by Tsai~\cite{tsai1987versatile} to derive constraints on the camera pose which are independent of the focal length and distortion parameters. For a 2D-3D correspondence, the idea is to only require that the 3D point projects onto the radial line passing through the 2D image point, and to ignore the radial offset. This constraint is called the Radial Alignment Constraint (RAC). This later gave rise to the 1D radial camera model (see \cite{thirthala2012radial}) which re-interprets the camera as projecting 3D points onto radial lines instead of 2D points. Since forward motion also moves the projections radially, it is not possible to estimate the forward translation using these constraints. In practice, the 1D radial camera model turns out to be equivalent to only considering the top two rows of the normal projection matrix. 
Instead of reprojection error, radial reprojection error measures the distance from 2D point to projected radial line, which is invariant to focal length and radial distortion parameters.

These ideas have also been applied to absolute pose estimation with unknown radial distortion. In Kukelova et al.~\cite{kukelova2013real}, the authors present a two-stage approach which first estimates the camera pose up to an unknown forward translation using the RAC. In a second step the last translation component is jointly estimated with the focal length and distortion parameters. This was later extended in Larsson et al.~\cite{larsson2019revisiting}. Camposeco et al.~\cite{camposeco2015non} applied a similar approach to non-parametric distortion models.

In this paper we take a similar approach as \cite{kukelova2013real,larsson2019revisiting,camposeco2015non}. However, instead of using just one frame, we can leverage multiple frames for the upgrade step since we consider multi-camera systems. We show it is possible to use joint poses of multiple (non-parallel) 1D radial cameras to transform the frames into the camera coordinate system for each single camera.


\section{Multi-Camera Calibration}

\begin{figure}[t]
    \begin{center}
    \includegraphics[trim=0 0 0 0,clip,width=1\linewidth]{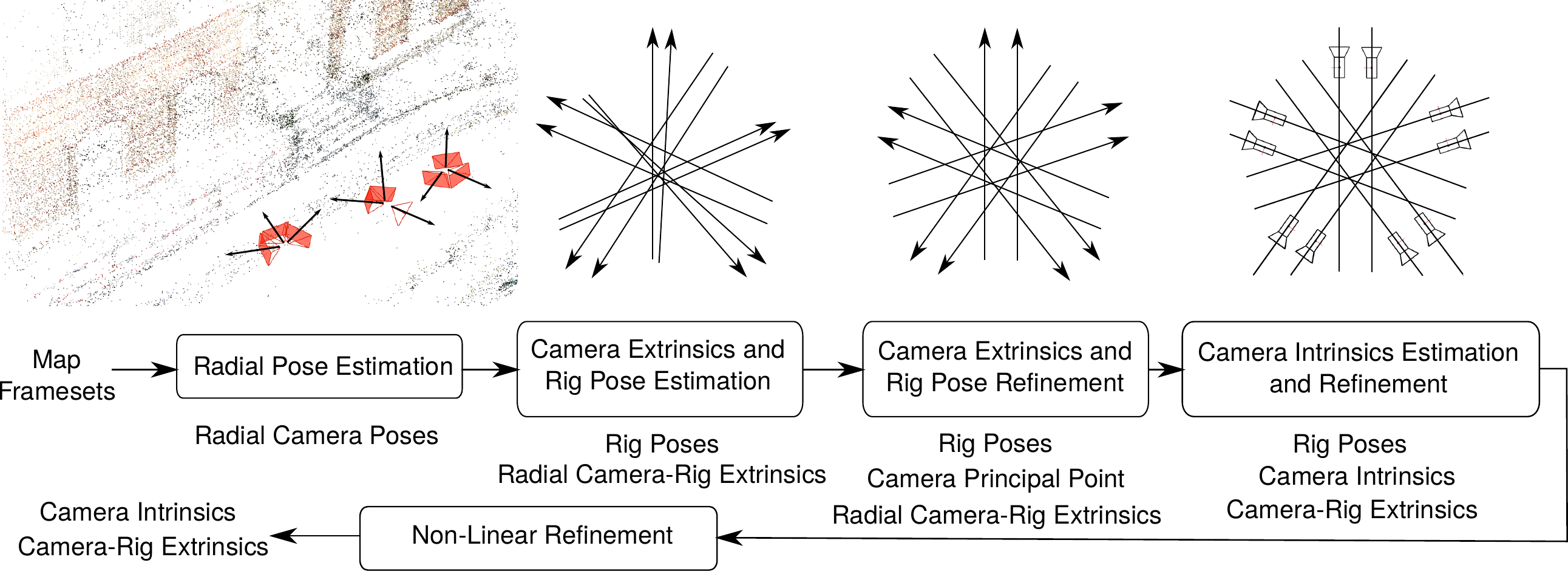}
    \end{center}
    \caption{\textbf{Illustration of the calibration pipeline.} The output of each step is placed below each block. In the first step we independently estimate the 1D radial pose of each camera. Next we robustly fit a rig to the estimated poses.  We refine the rig extrinsic parameters and poses by minimizing the radial reprojection errors. Then we upgrade each camera by estimating the last translation component jointly with the internal calibration. Finally we refine all parameters by minimizing the reprojection error.}
    \label{fig:pipeline}
\end{figure}

Now we present our framework for calibration of a multi-camera system. Our approach is similar to the infrastructure-based calibration method from Heng et al.~\cite{heng2015leveraging}. We improve on their approach in the following aspects:
\begin{itemize}
    \item We leverage state-of-the-art absolute pose solvers \cite{kukelova2013real,larsson2019revisiting} to also perform estimation of the camera intrinsic parameters, thus completely removing any need for pre-calibrating each camera independently. 
    \item We present a new robust estimation scheme to initialize the rig extrinsic parameters. Our experiments show that this greatly improves the robustness of the calibration method, especially on datasets with shorter image sequences.
    \item Finally we show it is possible to first partially estimate the rig extrinsics and pose before recovering the camera intrinsic parameters. This partial extrinsic knowledge allows us to more easily incorporate information from multiple images into the estimation.
\end{itemize}
Similarly to Heng et al.~\cite{heng2015leveraging} we assume that we have a sparse map of the environment. The input to our method is then a synchronized image sequence captured by the multi-camera system as it moves around in the mapped environment. The main steps of our pipeline are presented below and detailed in the next sections.
\begin{enumerate}
    \item \textbf{Independent 1D radial pose estimation.} We independently estimate a 1D radial camera pose (see Section~\ref{sec:radialpose}) for each image using RANSAC~\cite{fischler1981random}. 
    \item \textbf{Radial camera rig initialization.} We robustly fit a multi-camera rig with the 1D radial camera model to the estimated individual camera poses. 
    \item \textbf{Radial Bundle Adjustment.} We optimize the partial rig extrinsics and poses by minimizing the radial reprojection error (see Section~\ref{sec:radialpose}). Here we additionally refine the principal point for each camera.
    \item \textbf{Forward translation and intrinsic estimation.} 
    Using the partially known extrinsic parameters and poses of the rig we can transform all 2D-3D correspondences into the camera coordinate system (up to the unknown forward translation). This allows us to use the entire image sequence when initializing the intrinsic parameters and the forward translations~\cite{camposeco2015non}. 
    \item \textbf{Final refinement.} Finally, we perform bundle adjustment over rig poses, rig extrinsic and intrinsic parameters, minimizing the reprojection error over the entire sequence. 
\end{enumerate}
The entire calibration pipeline is illustrated in Figure~\ref{fig:pipeline}.

\subsection{The Sparse Map and Input Framesets}
One of the inputs to our method is a pre-built sparse feature map, which can be built using a standard Structure-from-Motion pipeline, e.g. COLMAP~\cite{schonberger2016structure}.
It is necessary to build a high-accuracy map in order to produce accurate calibration result. 
The map can be used as long as there is no large change in the environment.
The correct scale of the map can be derived either from a calibrated multi-camera system, e.g. stereo system, or by the user measuring some distances in both the real world and the map and scaling the map accordingly, e.g. by using a checkerboard.
In addition, a sequence of synchronized images captured in the map is recorded as the calibration dataset.
We define a frameset to be a set of images captured at the same timestamp from all different cameras.

\subsection{Initial Camera Pose Estimation} \label{sec:initial}
The first step of our pipeline is to independently estimate the pose of each image with respect to the pre-built map. Using the 1D radial camera model allows us to estimate the pose of the camera (up to forward translation) without knowing the camera intrinsic parameters (see Section~\ref{sec:radialpose}).
Similarly to Heng et al.~\cite{heng2015leveraging}, to find 2D-3D correspondences between the query image and the map we use a bag-of-words-based image retrieval against the mapping images, followed by 2D-2D image matching. For local features/descriptors we use upright SIFT~\cite{lowe2004distinctive}, but any local feature could be used. Once the putative 2D-3D correspondences are found we use the minimal solver from Kukelova et al.~\cite{kukelova2013real} (see Section~\ref{sec:radialpose}) in RANSAC to estimate the 1D radial camera pose. The principal point for each camera is initialized to the image center, a valid assumption for common cameras, and could be recovered accurately in later steps. Note that this only estimates the orientation and two components of the camera translation. At this stage we filter out any camera poses with too few inliers.

Alternatively we can also use the solvers from \cite{kukelova2013real,larsson2019revisiting} which directly solve for the intrinsic parameters. However, estimating the intrinsic parameters from a single image turns out to be significantly less stable. See Section~\ref{sec:eval_single_image_vs_merged} for a comparison of the errors in the intrinsic calibration when we perform the intrinsic calibration at this stage of the pipeline.


\subsection{Camera Extrinsics and Rig Poses Estimation} \label{sec:robustavg}
In the previous step we estimated the absolute poses for each image independently. Since we used the 1D radial camera model we only recovered the pose up to an unknown forward translation, i.e.~we estimated 
\begin{equation}
    T_{ij} = 
    \begin{bmatrix}
        {\large ~~R~~} & {\small \pmat{t_x\\t_y\\?}}
    \end{bmatrix} \enspace ,
\end{equation}
which transforms from the map coordinate system to the coordinate system of $i$th camera in the $j$th frameset. The goal now is to use the initial estimates to recover both the rig extrinsic parameters as well as the rig pose for each frameset in a robust way. 
In \cite{heng2015leveraging}, they simplify this problem by assuming that there is at least one frameset where each camera was able to get a pose estimate.
In our experiments this assumption was often not satisfied for shorter image sequences, leading to the method completely failing to initialize.

Let $P_i$ be the transform from the rig-centric coordinate system to the $i$th camera and let $Q_j$ be the transform from the map coordinate system to the rig-centric coordinate system for the $j$th frameset. A rig-centric coordinate system can be set to any rig-fixed coordinate frame since we only consider the relative extrinsics. In our case, it is set initially to be the same as the first camera with the unknown forward translation being zero. For noise-free measurements we should have
\begin{equation}
    T_{ij} = P_i Q_j, \quad (i,j) \in \Omega \enspace ,
\end{equation}
where $\Omega$ is the set of images that were successfully estimated in the previous step, i.e. $(i,j) \in \Omega$ if camera $i$ in frameset $j$ was successfully registered. 
Since we did not estimate the third translation component of $T_{ij}$, we restrict ourselves to finding the first two rows of the camera matrices, i.e.
\begin{equation}
        \hat{T}_{ij} = \hat{P}_i Q_j, \quad (i,j) \in \Omega \enspace ,
\end{equation}
where $\hat{T}_{ij}$ denotes the first two rows of $T_{ij}$ and similarly for $\hat{P}_i$. As described in Section~\ref{sec:radialpose} we can interpret $\hat{P}_i$ as 1D radial camera poses.  
If some $\hat{P}_i$ are known, then the rig poses $Q_j$ can be found by solving
\begin{equation}
    \bmat{\hat{T}_{1j} \\ \hat{T}_{2j} \\ \vdots} = \bmat{\hat{P}_{1} \\ \hat{P}_{2} \\ \vdots} Q_j \quad \text{where} \quad Q_j = \bmat{R & \vec{t} \\ \vec{0}^T & 1} \enspace ,
\end{equation}
which has a closed form solution using SVD~\cite{sorkine2017least}. Note that this requires that at least two cameras have non-parallel principal axes. We discuss this limitation more in Section~\ref{sec:limitations}. In turn, if the rig poses $Q_j$ are known, we can easily recover the rig extrinsic parameters as $\hat{P}_i = \hat{T}_{ij}Q_j^{-1}$.

To robustly fit the rig extrinsics $\hat{P}_i$ and rig poses $Q_j$ to the estimated absolute poses $\hat{T}_{ij}$, we solve the following minimization problem,
\begin{equation} \label{eq:robust_avg}
    \min_{\{\hat{P}_i\},\{Q_j\}} \sum_{(i,j) \in \Omega} \rho\left( d\left(\hat{T}_{ij},~\hat{P}_i Q_j\right) \right) \enspace ,
\end{equation}
where $\rho$ is a robust loss function and $d$ is a weighted sum of the rotation and translation errors.
Since this is a non-convex problem we perform a robust initialization scheme based on a greedy assignment in RANSAC.

In our case we randomly select any frameset with at least two cameras as initialization and assign the corresponding $\hat{P}_i$ using the relative poses from this frameset. Note that this might leave some $\hat{P}_i$ unassigned. We then use these assigned poses to estimate the rig poses $Q_j$ of any other frameset which also contains the already assigned $\hat{P}_i$. We can then iterate between assigning any of the missing $\hat{P}_i$ and estimating new $Q_j$. This back-and-forth search repeats until all of the rig extrinsics and rig poses are assigned. We repeat the entire process multiple times in a RANSAC-style fashion, keeping track of the best assignment with minimal radial reprojection over all frames. Finally, for the best solution we perform local optimization of \eqref{eq:robust_avg} using Levenberg-Marquardt.

This approach is similar to the rotation averaging methods in \cite{govindu2006robustness,olsson2011stable} which repeatedly build random minimum spanning trees in the pose-graph and assigns the absolute rotations based on these.

\subsection{Camera Extrinsics and Rig Poses Refinement}

We further refine the camera rig extrinsics and rig poses by performing bundle adjustment to minimize the radial reprojection error. In this step we also optimize over the principal point for each camera which was initialized to the image center. 
Let $\vec{X}_p$ be a 3D point and $\vec{x}_{ijp}$ its observation in the $i$th camera in frameset $j$. Then we optimize 
\begin{equation}
    \min_{\hat{P}_i,Q_j, \vec{c}_i} ~~\sum_{i,j,p}\rho\left(\left\|\pi_r\left(\hat{P}_iQ_j\vec{X}_p,~\vec{x}_{ijp}-\vec{c}_i\right) - (\vec{x}_{ijp}-\vec{c}_i)\right\|^2\right) \enspace ,
\end{equation}
where $\rho$ is a robust loss function, $\vec{c}_i$ is the principal point of the $i$th camera and $\pi_r : \mathbb{R}^2 \times \mathbb{R}^2 \to \mathbb{R}^2$ is the orthogonal projection of the second argument onto the line generated by the first, i.e. 
$\pi_r(\vec{u},\vec{v}) = \frac{\vec{u}^T\vec{v}}{\vec{u}^T\vec{u}}\vec{u}.$



\subsection{Camera Upgrading and Refinement}
In this step, we estimate the internal calibration as well as the remaining unknown translation component for each camera. By transforming all 2D-3D correspondences into the rig frame, we can leverage data from all framesets.

From the previous step we have estimated the camera rig extrinsics $P_i$, except for the third component of the translation vector, i.e. $t_{z,i}$. The 3D points mapped into each camera's coordinate system can then be written as
\begin{equation}
    \vec{Z}_{ijp} + t_{z,i}\vec{e}_z = P_iQ_j\vec{X}_p \quad \text{where} \quad \vec{e}_z = \pmat{0,0,1}^T.
\end{equation}
Now we can use the minimal solvers from Kukelova et al.~\cite{kukelova2013real} and Larsson et al.~\cite{larsson2019revisiting} for jointly estimating $t_{z,i}$ and the intrinsic parameters. 
To further remove outlier correspondences, we again use RANSAC to robustly initialize the parameters.
Additionally, we perform non-linear optimization to refine the intrinsics and $t_{z,i}$ by minimizing the reprojection error as
\begin{equation}
        \min_{\theta_i, t_{z}} \sum_{j,p}\rho\left(\left\|{\pi}_{\theta_i}\left(\vec{Z}_{ijp} + t_{z,i}\vec{e}_z\right) - \vec{x}_{ijp}\right\|^2\right) \enspace ,
\end{equation}
where $\theta_i$ are the intrinsic parameters and $\pi_{\theta_i}$ denotes the projection into image space. Note that here we use full distortion model instead of pure radial distortion. This is done for each camera individually.

\subsection{Final Refinement} \label{sec:finalba}
In the final step, we optimize all the camera intrinsics, extrinsics and rig poses by minimizing the reprojection error. The optimization problem is 
\begin{equation}
        \min_{P_i,Q_j, \theta_i}\sum_{i,j,p}\rho\left(\left\|\pi_{\theta_i}(P_iQ_j\vec{X}_p) - \vec{x}_{ijp}\right\|^2\right) \enspace .
\end{equation}
Optionally the 3D scene points can be added into optimization problem, in case the scene points are not accurate enough.

\section{Implementation}
\label{sec:limitations}
Our implementation is based on the infrastructure-based calibration from the CamOdoCal library~\cite{heng2013camodocal}. The sparse map is built by COLMAP~\cite{schonberger2016structure}, which uses upright SIFT~\cite{lowe2004distinctive} features and descriptors. For the camera model, pinhole with radial-tangential distortion and pinhole with equidistant distortion~\cite{kannala2006generic} are supported to suit different cameras. The optimization is solved with the Levenberg-Marquardt algorithm using the Ceres Library~\cite{agarwal2018ceres} and we use the Cauchy loss with scale parameter 1 as the robust loss function.

\PAR{Limitations.} 
Note that to robustly fit the rig extrinsics among different framesets requires that at least two cameras in the rig have non-parallel principal axes, otherwise Equation~\ref{eq:robust_avg} fails to determine the rig pose. 
However, camera rigs with parallel principal axes, usually stereo camera setups, can be easily calibrated through existing calibration methods. 
Other cases, \eg two cameras with opposite direction, commonly equipped in mobile phones, can be calibrated by our proposed calibration variant \textbf{Inf+RD+RA} described in Section~\ref{sec:evaluation-setup}, which uses pose solvers that can estimate both the poses and intrinsics per frame.

\section{Experimental Evaluation}
For the experimental evaluation of our method we first consider two different multi-camera systems, one pentagonal camera rig with ten cameras arranged in five stereo pairs (Figure~\ref{fig:rig-image}a) and a ski helmet with five GoPro Hero7 Black cameras attached (Figure~\ref{fig:rig-image}b). For the GoPro cameras we record in wide \gls{fov} mode, which roughly corresponds to 120$^\circ$ degree horizontal \gls{fov}. The cameras on the pentagonal rig have circa 70$^\circ$ horizontal \gls{fov}.

\subsection{Evaluation Datasets and Setup}
\label{sec:evaluation-setup}
\begin{figure}[tbp]
\centering
\includegraphics[width=0.24\textwidth]{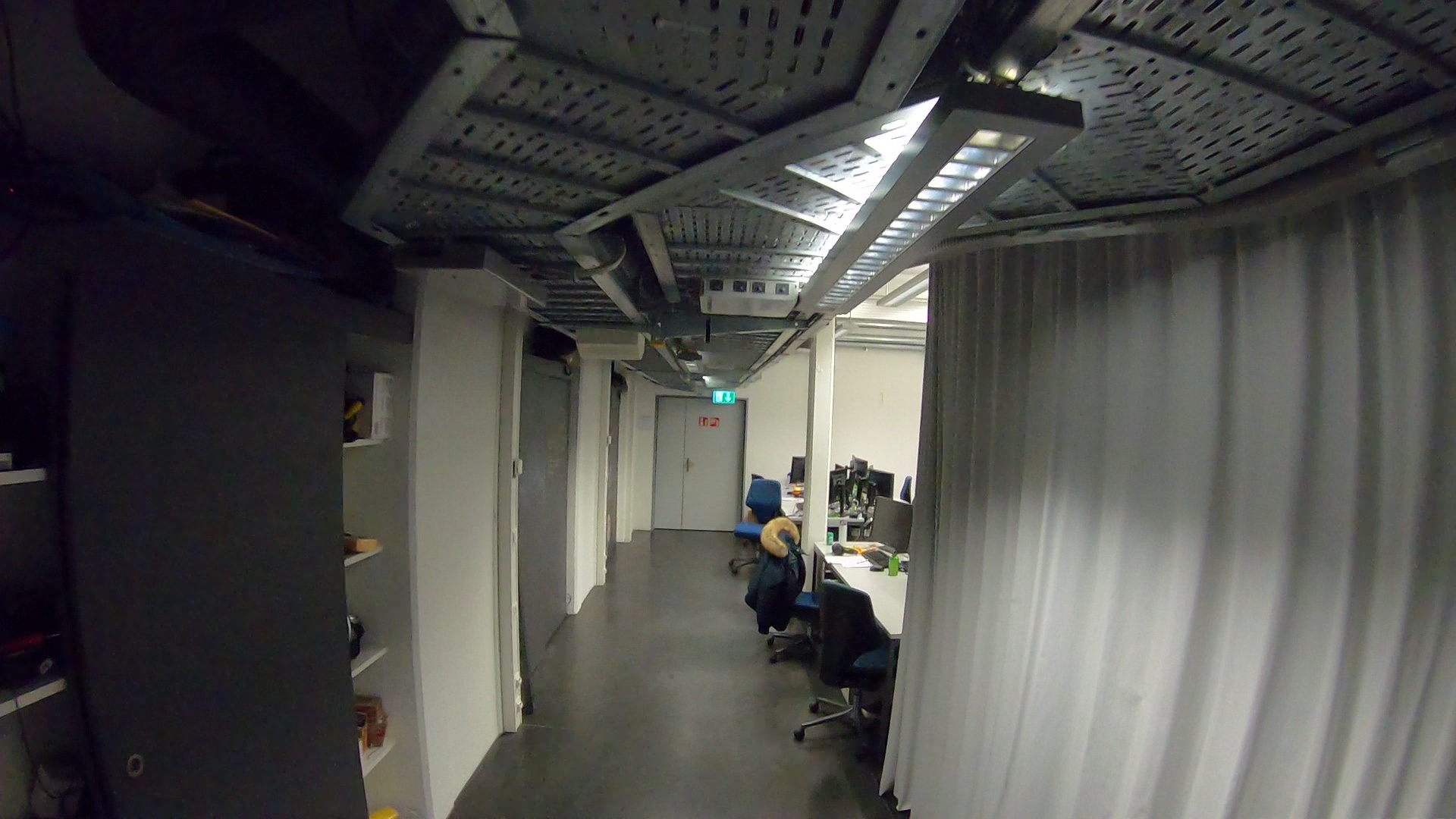}
\includegraphics[width=0.24\textwidth]{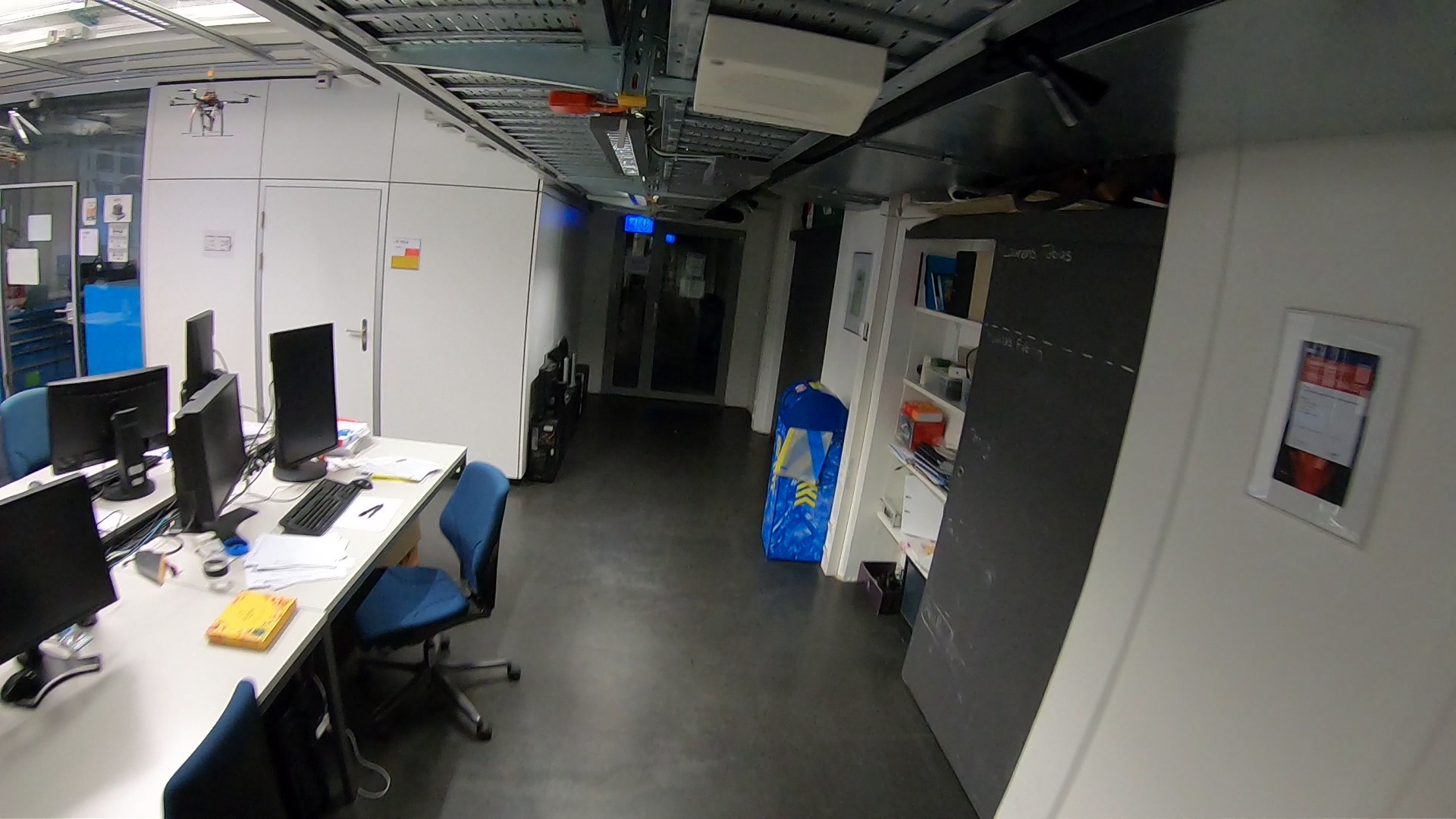}
\includegraphics[width=0.24\textwidth]{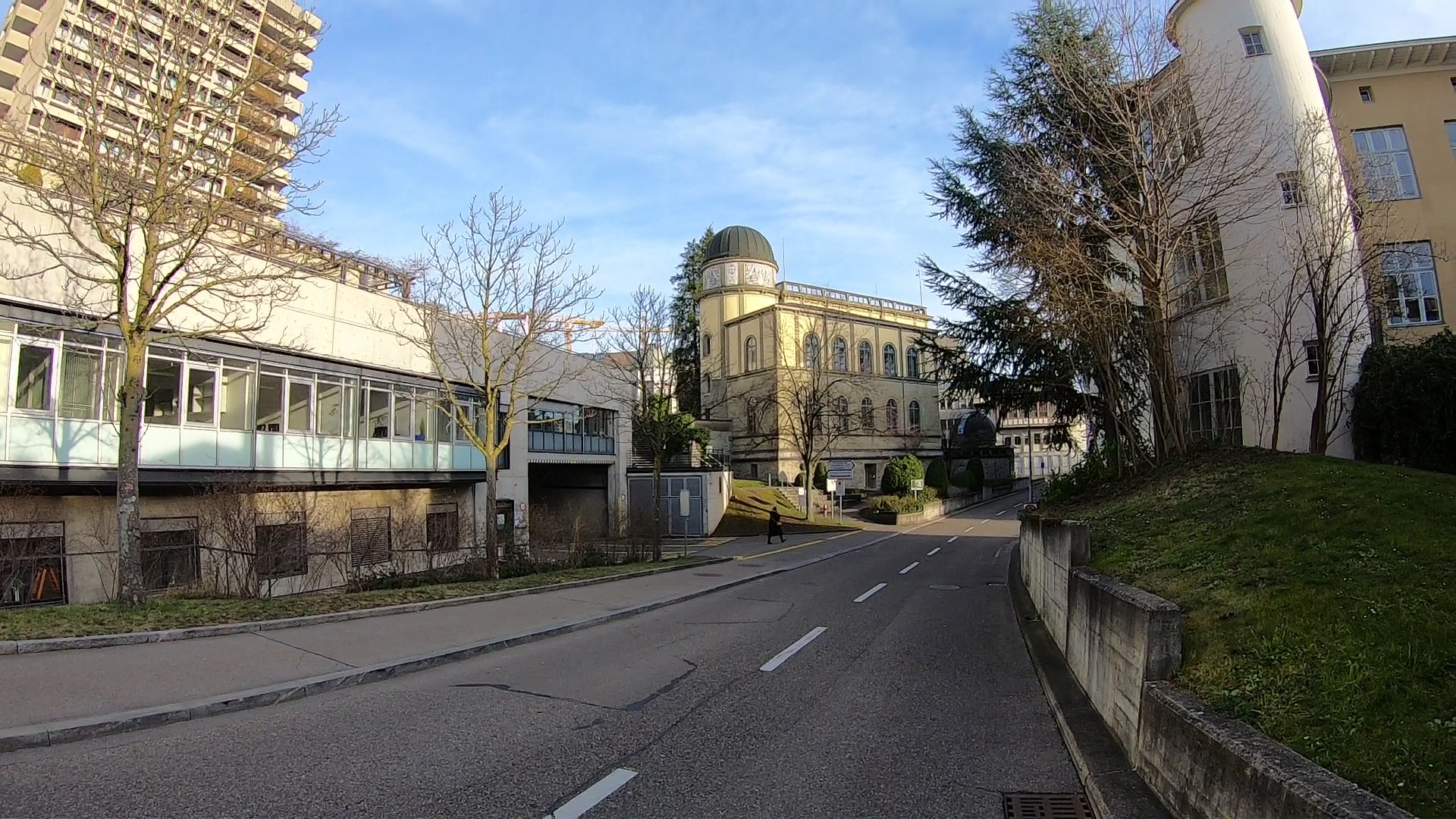}
\includegraphics[width=0.24\textwidth]{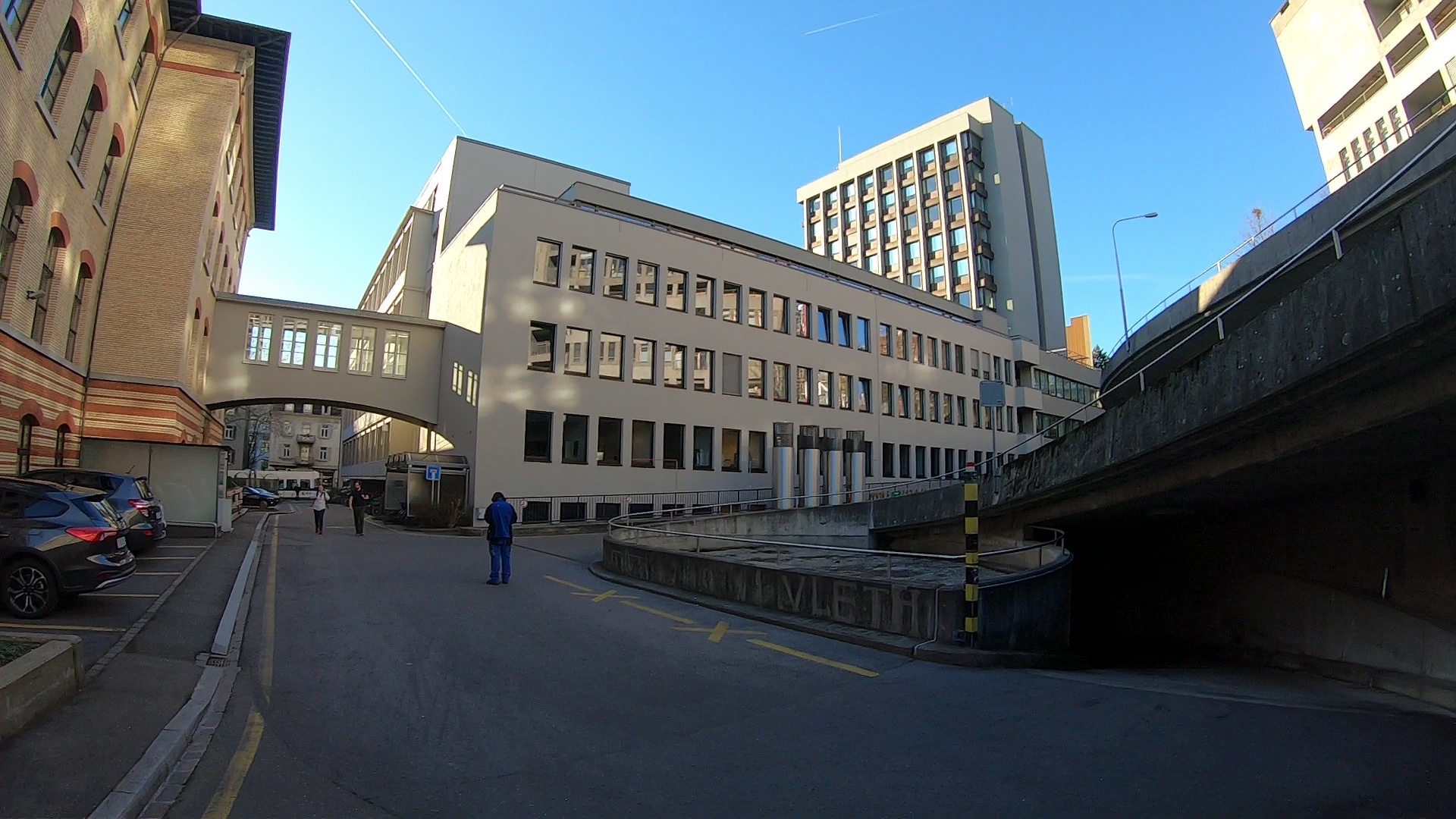}
\caption{\textbf{Sample images of the environment}. \textit{Left:} Indoor environment in a lab room. \textit{Right:} Outdoor environment on an urban road.}
\label{fig:env-image}
\end{figure}

To validate our method we record datasets in both indoor and outdoor environments. See Figure~\ref{fig:env-image} for example images. For each dataset we record
    a mapping sequence with the GoPro Hero Black 7 in linear mode\footnote{Linear mode provides in-camera undistorted images with a reduced \gls{fov}.}, 
    calibration sequences with both the pentagonal rig and the GoPro helmet, 
    and Aprilgrid sequences to allow for offline calibration and validation. 
We use the calibration toolbox Kalibr~\cite{maye2013self} on the Aprilgrid datasets to create a \textit{ground-truth} calibration for comparison. As far as we know there is no competing method that performs infrastructure-based multi-camera calibration with unknown intrinsic parameters. 
We augment the original pipeline from Heng et al.~\cite{heng2015leveraging} with radial distortion solvers from Larsson et al.~\cite{larsson2019revisiting} as candidates to join the comparison.
In particular, we compare the following  approaches:
\begin{itemize}
    \item \textbf{Inf+K.} The infrastructure-based method from Heng et al.~\cite{heng2015leveraging}.
    \item \textbf{Inf+K+RI.} Same as Inf+K but with refinement of the intrinsic parameters during the final bundle adjustment.
    \item \textbf{Inf+RD.} We replace the P3P solver in \cite{heng2015leveraging} with the pose solvers from Larsson~et~al.~\cite{larsson2019revisiting} which also estimate distortion parameters and focal length.
    \item \textbf{Inf+RD+RA.} We add a robust rig averaging similar to Section~\ref{sec:robustavg}.
    \item \textbf{Inf+1DR+RA.} The proposed pipeline as described in Section~\ref{sec:initial}-\ref{sec:finalba} which delays estimation of the intrinsic parameters using 1D radial cameras.
\end{itemize}
Note that \textbf{Inf+K} and \textbf{Inf+K+RI} use the intrinsic parameters from running Kalibr on the Aprilgrid images and join the competition as references. To evaluate the resulting calibrations we robustly align the estimated camera rigs with the camera rigs obtained from Kalibr~\cite{maye2013self} and compute the difference in the rotations (degrees) and camera centers (centimetres). To evaluate the intrinsic parameters we validate the calibration on the Aprilgrid datasets and report the average reprojection error (pixels).

\subsection{Calibration Accuracy and Run-Time on Full Image Sequence}
\begin{figure}[tbp]
\centering
\includegraphics[width=0.32\textwidth,height=0.17\textwidth]{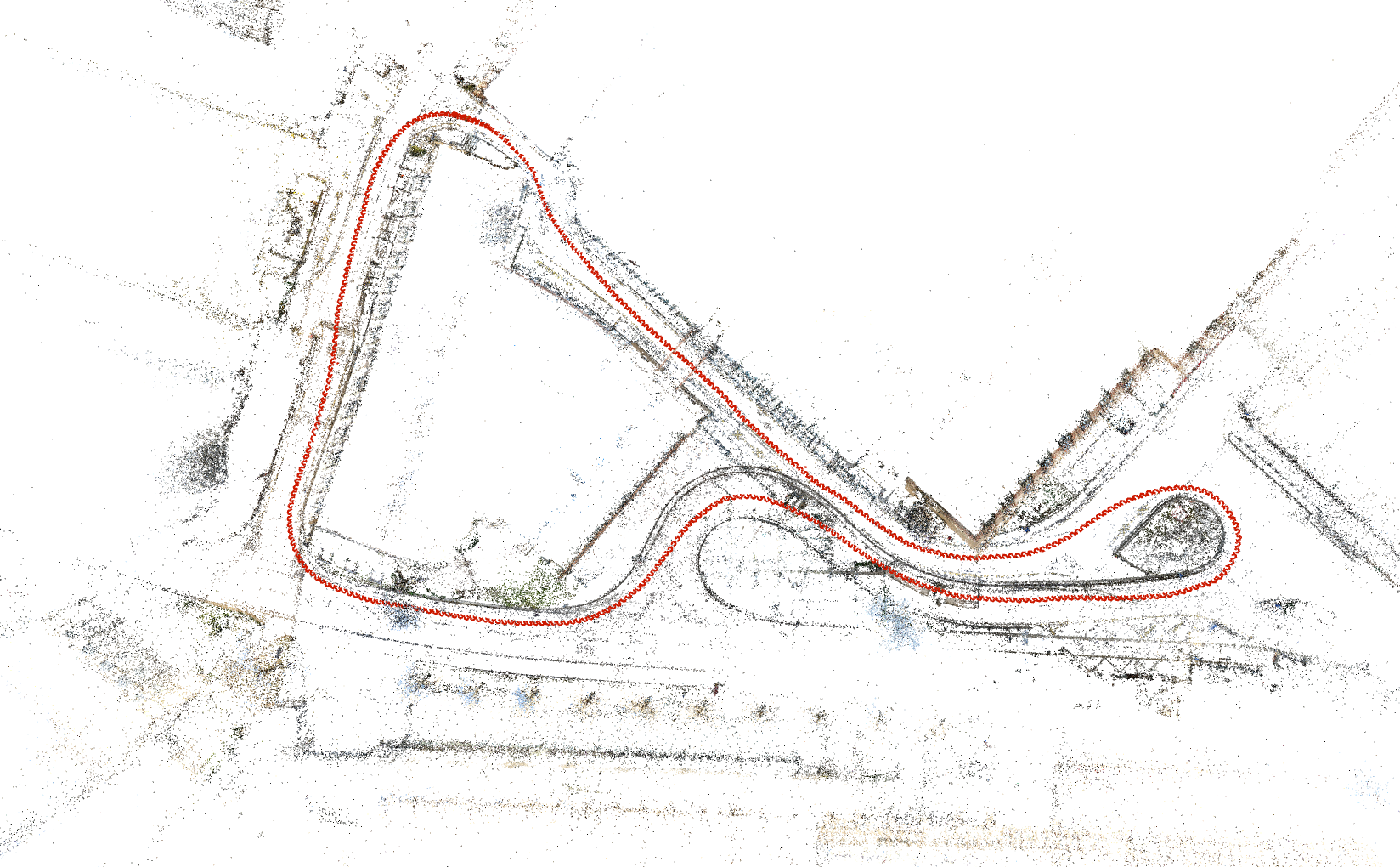}
\includegraphics[width=0.32\textwidth,height=0.17\textwidth]{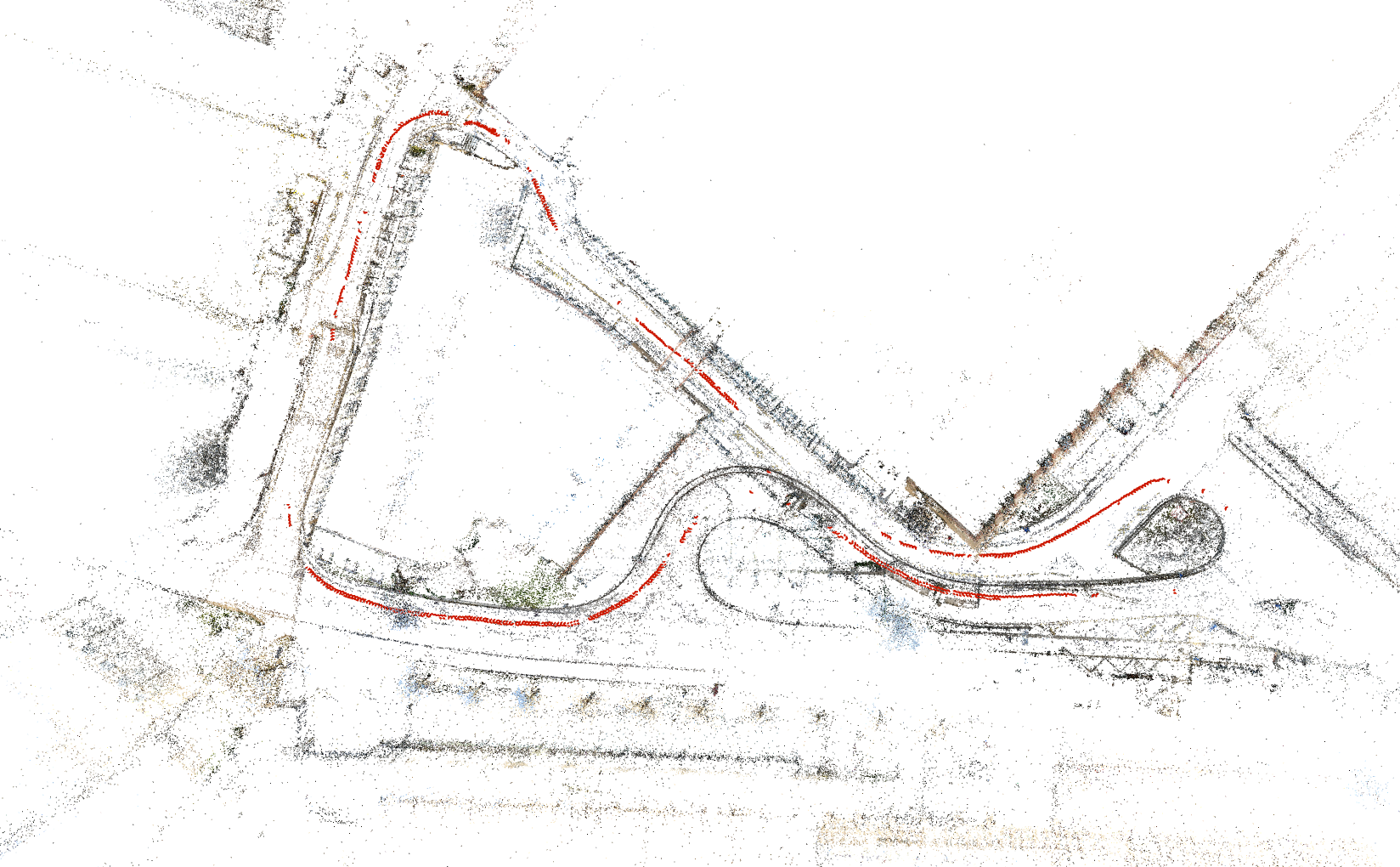}
\includegraphics[width=0.32\textwidth,height=0.17\textwidth]{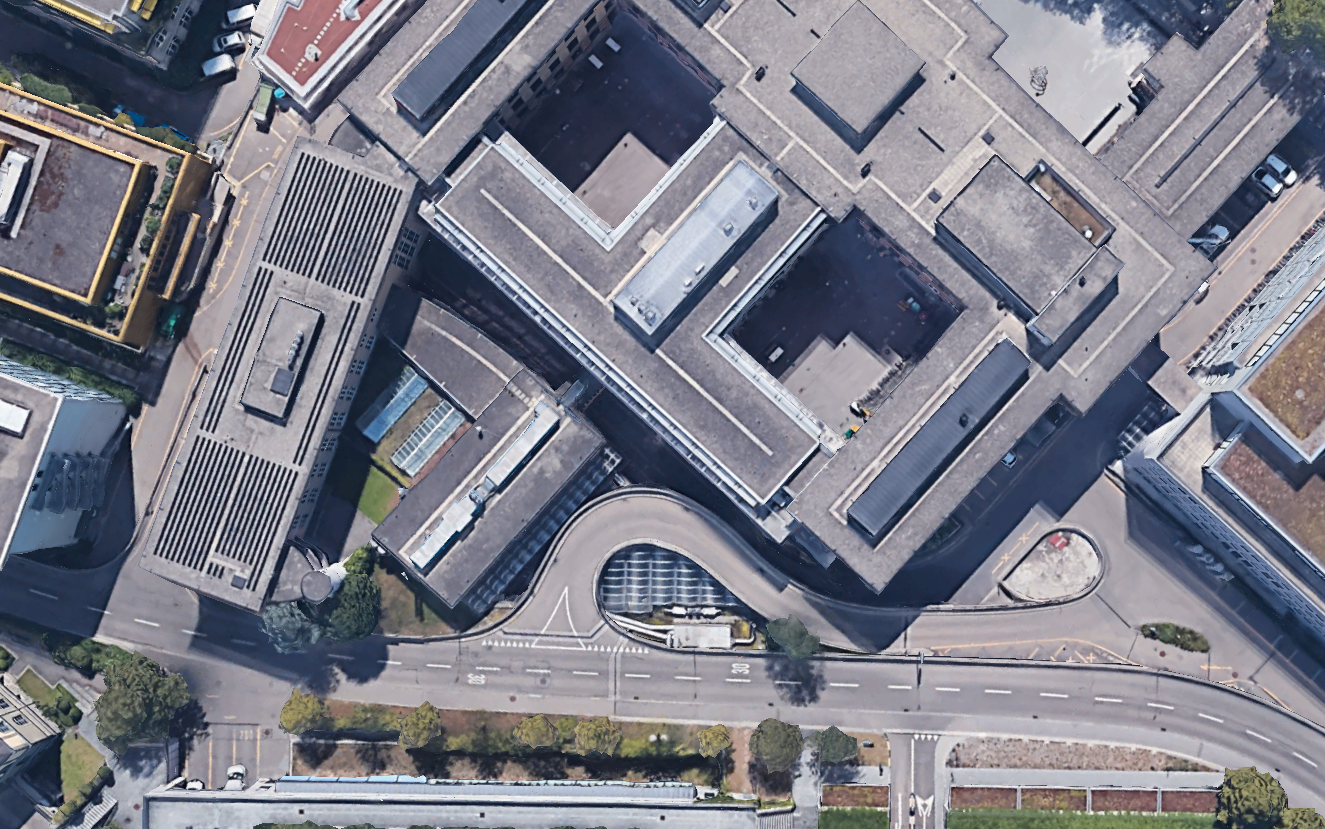}
\caption{\textbf{Experiments in outdoor urban environment}. \textit{Left:} The sparse reconstruction from COLMAP~\cite{schonberger2016structure} with mapping sequence shown in red. \textit{Middle:} The same scene with frames used for calibration in red. \textit{Right:} Aerial view of the scene.}
\label{fig:outdoor-experiment}
\end{figure}

First we aim to evaluate the accuracy of the calibrations by running the methods on the entire calibration sequences. See Figure~\ref{fig:outdoor-experiment} for a visualization of camera poses recovered in the \textit{Outdoor} dataset. The results can be found in Table~\ref{tbl:accuracy}.  We can see that, using infrastructure-based calibration methods, we are able to obtain similar quality results as classical Aprilgrid based methods. In this case, the three methods \textbf{Inf$+$RD},~ \textbf{Inf$+$RD$+$RA}, and \textbf{Inf$+$1DR$+$RA} all had very similar performance. Note also that the ground truth we are comparing to is not necessarily perfect. In practice, we find that with similar datasets recorded at the same time, the extrinsic results differ up to 0.3$^\circ$ and 0.5 centimeters.

For run-time, we run our method on a DELL Laptop equipped with 16 GB RAM, an i7-9750H CPU and a GTX1050 GPU. A comparison of the processing time of each pipeline is shown in Table~\ref{tab:runtime}. Our method \textbf{Inf$+$1DR$+$RA} takes a similar amount of time while removing the need for pre-calibration required by \textbf{Inf$+$K$+$RI}, and runs much faster than the pattern-based method \textbf{Kalibr}.


\begin{table}[t]
\caption{\textbf{Evaluation of calibration accuracy} The errors are with respect to the calibration obtained from the Aprilgrid datasets with Kalibr~\cite{maye2013self}. Note that \textbf{Inf$+$K} and \textbf{Inf$+$K$+$RI} use the ground-truth intrinsic parameters as input.}
\centering
\resizebox{0.8\textwidth}{!}{%
\begin{tabularx}{\textwidth}{ll *{5}{>{\centering\arraybackslash}X}} 
                  \hfill \textbf{Inf$+$} && {\small \textbf{K}} & {\small \textbf{K$+$RI}}  & {\small \textbf{RD}}  & {\small \textbf{RD$+$RA}} &{\small \textbf{1DR$+$RA}} \\ \toprule
    \textbf{GoPro Helmet / Indoor}               &~&  &&&& \\

     Reproj. error (px) && 0.283 & 0.270 & 0.526 &0.412 &  0.270 \\
     Rot. error (degree) &&   0.193 & 0.320 & 0.328 &0.319 & 0.321 \\
     Trans. error (cm) &&   0.780 & 0.418 & 0.430 &0.435 & 0.426 \\ \midrule
    \textbf{GoPro Helmet / Outdoor}               &~&  &&&& \\
       
     Reproj. error (px) && 0.339 & 0.337 &0.337 & 0.336& 0.337  \\
     Rot. error (degree) &&  0.141 & 0.188 & 0.187 &0.187 & 0.187 \\
     Trans. error (cm) &&   0.642 & 0.392 &0.385 & 0.390 &0.384  \\ \midrule
         \textbf{Pentagonal / Indoor }              &~& &&&& \\
       
     Reproj. error (px) && 0.230 & 0.281 & 0.280 &0.308 & 0.282 \\
     Rot. error (degree) &&  0.293 & 0.548 & 0.545 &0.543 & 0.543 \\
     Trans. error (cm) &&   1.316 & 0.366 & 0.372 &0.377 & 0.376 \\\midrule
         \textbf{Pentagonal / Outdoor}               &~&  &&&& \\
        
     Reproj. error (px) && 0.198 & 0.268 & 0.268 &0.263 & 0.271 \\
     Rot. error (degree) &&  0.295 & 0.570 & 0.566 &0.568 & 0.567 \\
     Trans. error (cm) &&   2.217 & 0.441 & 0.419 &0.417 & 0.423 \\ \bottomrule
\end{tabularx}
}
\label{tbl:accuracy}
\end{table}

\begin{table}[t]
\centering
\caption{\textbf{Run-Time Comparison.} Table lists the average runtime (in minutes) for different methods on calibration sequences with 500 framesets. The runtime for Inf+K+RI and Inf+1DR+RA consists of indoor/outdoor cases.}
\label{tab:runtime}
\resizebox{0.7\textwidth}{!}{%

    \begin{tabularx}{0.8\textwidth}{ll *{3}{>{\centering\arraybackslash}X}} 
Runtime (min) && \textbf{Inf+K+RI} & \textbf{Inf+1DR+RA} & \textbf{Kalibr} \\ \toprule
GoPro Helmet  && 9.5 / 11.3                 & 10.9 / 12.2                  & 24.5   \\ \hline
Pentagonal    && 7.0 / 9.6                  & 11.0 / 15.4                  & 113.0  \\ \bottomrule
\end{tabularx}
}
\end{table}

\subsection{Evaluation of Robustness on Shorter Image Sequences}
\label{sec:robust}
In the previous section we saw that if we have enough data we are able to achieve high-quality calibration results. In this section we instead evaluate the robustness of the method when input data is more limited. For many applications this is an important scenario since you might want to find the camera calibration as quickly as possible to enable other tasks which depend on knowing the camera calibration. To perform the experiment we select multiple sub-sequences and try to calibrate from these. For each sequence we select 10 framesets which approximately differ by one second (the datasets were captured at normal walking speed). Table~\ref{tbl:robustness} shows the percentage of frames where the calibration-methods were able to calibrate the complete rig, as well as the percentage of sequences which gave good calibrations (defined as rotation error below 1 degree and translation below 1 cm for indoor and 2 cm for outdoor). The total number of sequences were 313 (penta) and 173 (GoPro). Table~\ref{tbl:robustness} shows the superior robustness of our approach.


\begin{table}[t]
    \caption{\textbf{Comparison of robustness for shorter image sequences.} Table shows the percentage of sequences which we were able to estimate a complete frameset and the percentage of sequences of sequences that were accurately calibrated. A good calibration is defined in Section~\ref{sec:robust}. }
    \label{tbl:robustness}
    \centering
    \resizebox{0.8\textwidth}{!}{%
    \begin{tabularx}{\textwidth}{c rl *{3}{>{\centering\arraybackslash}X}} 
        &\hfill \textbf{Inf$+$} && {\small \textbf{RD}}  & {\small \textbf{RD$+$RA}} &{\small \textbf{1DR$+$RA}} \\ \toprule
\multirow{2}{*}{GoPro Helmet / Indoor~}& Complete && 54.5 &\bf	98.3& \bf 98.3 \\
                              & Good     && 44.9 &	75.6 &\bf 	79.0  \\ \midrule
                              
\multirow{2}{*}{GoPro Helmet / Outdoor~}& Complete && 67.6 &	97.7&\bf 	98.3  \\
                              & Good     && 38.1 & 45.5 & \bf 48.3  \\ \midrule
                              
\multirow{2}{*}{Pentagonal / Indoor~}& Complete && 31.9	& 68.4	& \bf 69.0 \\
                              & Good     && 23.0 & 43.1 & \bf 44.4 \\ \midrule
                              
\multirow{2}{*}{Pentagonal / Outdoor~}& Complete && 28.8 & 79.2 & \bf 80.5  \\
                              & Good     && 21.1 & 38.3 &\bf  41.5 \\ \bottomrule \\
    \end{tabularx}
    }

\end{table}

\subsection{Evaluation of Initial Estimates} \label{sec:eval_single_image_vs_merged}
\begin{figure}[tb]
    \centering
    \begin{minipage}{0.45\textwidth}
    \begin{tabular}{cc}
    \textbf{Inf$+$RD$+$RA} & \textbf{Inf$+$1DR$+$RA} \\
    \includegraphics[trim=150 100 120 80,clip,width=0.49\textwidth]{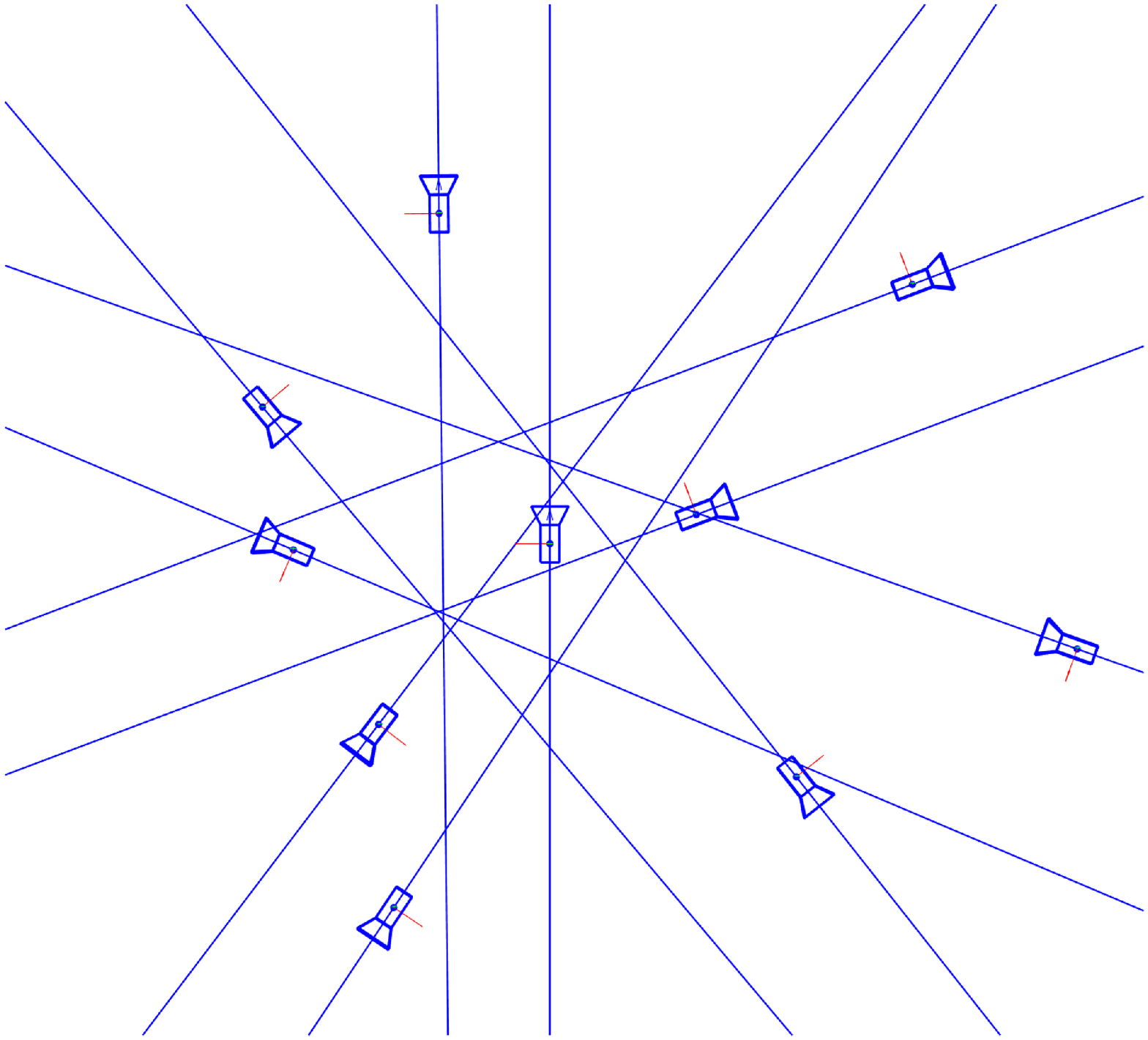}&
    \includegraphics[trim=150 100 120 80,clip,width=0.49\textwidth]{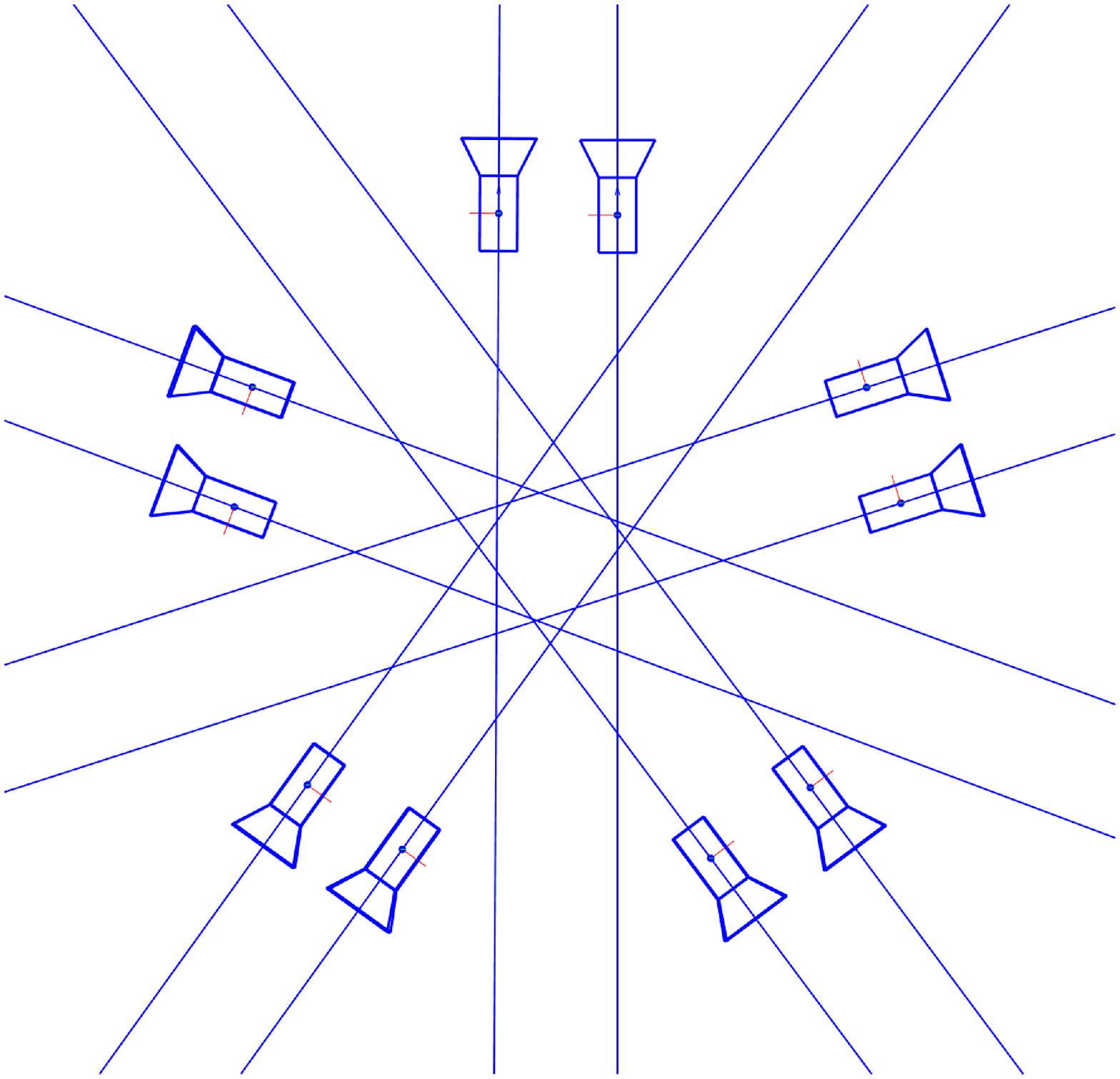}
    \end{tabular}
    \end{minipage}
    \begin{minipage}{0.24\textwidth}
%
%
\definecolor{mycolor1}{rgb}{0.00000,0.44700,0.74100}%
\definecolor{mycolor2}{rgb}{0.85000,0.32500,0.09800}%
\begin{tikzpicture}
\tikzstyle{every node}=[font=\small,scale=0.75]

\begin{axis}[%
width=\linewidth,
height=0.4\linewidth,
scale only axis,
xmin=0,
xmax=5,
ymin=0,
ymax=2.5,
axis lines=left,
xlabel={Rotation error \textit{(degree)}},
yticklabels={,,},
axis background/.style={fill=white},
legend style={legend cell align=left, align=left, draw=white!15!black, style={nodes={scale=0.8, transform shape}}}
]
\addplot [color=red,line width=1pt]
  table[row sep=crcr]{%
0	0.00015210995804485\\
0.0505050505050505	0.000862512235272665\\
0.101010101010101	0.00399307151662073\\
0.151515151515152	0.0151799771624164\\
0.202020202020202	0.0476995980319856\\
0.252525252525253	0.124827976657844\\
0.303030303030303	0.274461300588011\\
0.353535353535354	0.512441612626036\\
0.404040404040404	0.823351752376224\\
0.454545454545455	1.15726994375849\\
0.505050505050505	1.44909747842155\\
0.555555555555556	1.64239311347197\\
0.606060606060606	1.69962527816168\\
0.656565656565657	1.60860459230147\\
0.707070707070707	1.39839481297889\\
0.757575757575758	1.14285689141966\\
0.808080808080808	0.924292733118508\\
0.858585858585859	0.781745301214155\\
0.909090909090909	0.697973002955476\\
0.95959595959596	0.633589505331283\\
1.01010101010101	0.564755279829678\\
1.06060606060606	0.491866052983053\\
1.11111111111111	0.427166542714148\\
1.16161616161616	0.380159906442911\\
1.21212121212121	0.348708814570923\\
1.26262626262626	0.320036554735137\\
1.31313131313131	0.282092206956954\\
1.36363636363636	0.234297654175868\\
1.41414141414141	0.186145462204144\\
1.46464646464646	0.147261256109509\\
1.51515151515152	0.120926072013647\\
1.56565656565657	0.105010035494192\\
1.61616161616162	0.0958139106471732\\
1.66666666666667	0.0911239757084572\\
1.71717171717172	0.091259878934888\\
1.76767676767677	0.097054783879767\\
1.81818181818182	0.105966784138863\\
1.86868686868687	0.111138734430161\\
1.91919191919192	0.106060029869656\\
1.96969696969697	0.0899487607307494\\
2.02020202020202	0.0675806391391889\\
2.07070707070707	0.0449425243224085\\
2.12121212121212	0.0261684995307325\\
2.17171717171717	0.0130636365629048\\
2.22222222222222	0.00546154335644989\\
2.27272727272727	0.00187533035779386\\
2.32323232323232	0.000521775331317257\\
2.37373737373737	0.000116643538093831\\
2.42424242424242	2.08467794385402e-05\\
2.47474747474747	2.97011247924839e-06\\
2.52525252525253	3.36788428812933e-07\\
2.57575757575758	3.03664580980905e-08\\
2.62626262626263	2.1760033545092e-09\\
2.67676767676768	1.23887235773466e-10\\
2.72727272727273	5.60303182197328e-12\\
2.77777777777778	2.01283430407039e-13\\
2.82828282828283	5.74325114459188e-15\\
2.87878787878788	1.30154405445502e-16\\
2.92929292929293	2.34262827997831e-18\\
2.97979797979798	3.34879622793696e-20\\
3.03030303030303	3.99517963938743e-22\\
3.08080808080808	1.99489811800889e-21\\
3.13131313131313	1.63182003992792e-19\\
3.18181818181818	1.0632353839698e-17\\
3.23232323232323	5.50951543373203e-16\\
3.28282828282828	2.27090546913328e-14\\
3.33333333333333	7.44677802139018e-13\\
3.38383838383838	1.94316924187006e-11\\
3.43434343434343	4.0357274338463e-10\\
3.48484848484848	6.6727463230045e-09\\
3.53535353535354	8.78551712135083e-08\\
3.58585858585859	9.21346356357941e-07\\
3.63636363636364	7.69816911506624e-06\\
3.68686868686869	5.12599459140975e-05\\
3.73737373737374	0.000272089089825798\\
3.78787878787879	0.00115159226212486\\
3.83838383838384	0.00388728706298587\\
3.88888888888889	0.0104676847293441\\
3.93939393939394	0.0224902294398294\\
3.98989898989899	0.0385606036875732\\
4.04040404040404	0.0527654729365049\\
4.09090909090909	0.0576290382363335\\
4.14141414141414	0.0502373604922002\\
4.19191919191919	0.0349536983284084\\
4.24242424242424	0.0194091454445894\\
4.29292929292929	0.00860029113580119\\
4.34343434343434	0.00304048080774226\\
4.39393939393939	0.000857447681976156\\
4.44444444444444	0.00019284615193244\\
4.49494949494949	3.45815894730644e-05\\
4.54545454545455	4.943063039859e-06\\
4.5959595959596	5.63051338717517e-07\\
4.64646464646465	5.10955790095544e-08\\
4.6969696969697	3.69306210330129e-09\\
4.74747474747475	2.12542503564595e-10\\
4.7979797979798	9.73763954827612e-12\\
4.84848484848485	3.55065987316285e-13\\
4.8989898989899	1.03018991832206e-14\\
4.94949494949495	2.37788443066197e-16\\
5	4.36565029069154e-18\\
};
\addlegendentry{\textbf{Inf$+$RD$+$RA}}

\addplot [color=blue,line width=1pt]
  table[row sep=crcr]{%
0	0.00641044360534877\\
0.0505050505050505	0.0288156179827256\\
0.101010101010101	0.0951118380302141\\
0.151515151515152	0.253779767399315\\
0.202020202020202	0.578597325113216\\
0.252525252525253	1.10285172466517\\
0.303030303030303	1.68239416811352\\
0.353535353535354	2.04580214127358\\
0.404040404040404	2.09500161483437\\
0.454545454545455	1.9897394530545\\
0.505050505050505	1.8646965434623\\
0.555555555555556	1.67286845927213\\
0.606060606060606	1.37667915002703\\
0.656565656565657	1.06516925214021\\
0.707070707070707	0.796820499786326\\
0.757575757575758	0.55858707845605\\
0.808080808080808	0.372291228488396\\
0.858585858585859	0.271693632377348\\
0.909090909090909	0.233032229617816\\
0.95959595959596	0.202308621991375\\
1.01010101010101	0.152391351217701\\
1.06060606060606	0.0907493850012617\\
1.11111111111111	0.0395161530661905\\
1.16161616161616	0.0118141638273229\\
1.21212121212121	0.00238031924006259\\
1.26262626262626	0.000803606544255873\\
1.31313131313131	0.00306187423309068\\
1.36363636363636	0.0114165451766431\\
1.41414141414141	0.0269532792375241\\
1.46464646464646	0.0400441769919561\\
1.51515151515152	0.0379217791648032\\
1.56565656565657	0.0255987974828824\\
1.61616161616162	0.0213454492954356\\
1.66666666666667	0.0312433778240383\\
1.71717171717172	0.0409914155784322\\
1.76767676767677	0.0354733485432059\\
1.81818181818182	0.019388412717784\\
1.86868686868687	0.00666166744392515\\
1.91919191919192	0.00144451000004229\\
1.96969696969697	0.000284725515968302\\
2.02020202020202	0.000790030440641159\\
2.07070707070707	0.00419109602189318\\
2.12121212121212	0.0142773005938987\\
2.17171717171717	0.0307259834261962\\
2.22222222222222	0.0426839888055779\\
2.27272727272727	0.0422457933962739\\
2.32323232323232	0.0397457822350141\\
2.37373737373737	0.0444720339100613\\
2.42424242424242	0.0488317748820716\\
2.47474747474747	0.051030766295797\\
2.52525252525253	0.0618244615647554\\
2.57575757575758	0.0795747092076228\\
2.62626262626263	0.0898942794371781\\
2.67676767676768	0.0845601745516816\\
2.72727272727273	0.0642673996751181\\
2.77777777777778	0.0368951375112701\\
2.82828282828283	0.0148666651580201\\
2.87878787878788	0.00399243102318456\\
2.92929292929293	0.00069386699529517\\
2.97979797979798	7.68703925232455e-05\\
3.03030303030303	5.3881309013446e-06\\
3.08080808080808	2.38085777648881e-07\\
3.13131313131313	6.6203529905291e-09\\
3.18181818181818	1.15747775912645e-10\\
3.23232323232323	1.27189211870691e-12\\
3.28282828282828	8.78230153387703e-15\\
3.33333333333333	3.81018059448671e-17\\
3.38383838383838	1.03858492878822e-19\\
3.43434343434343	1.77864023009474e-22\\
3.48484848484848	1.91372362296474e-25\\
3.53535353535354	1.2936390140148e-28\\
3.58585858585859	5.49400051621733e-32\\
3.63636363636364	1.46590380675262e-35\\
3.68686868686869	2.45732831459497e-39\\
3.73737373737374	2.5879847441053e-43\\
3.78787878787879	1.71238334686997e-47\\
3.83838383838384	7.11837737433611e-52\\
3.88888888888889	1.85909565362278e-56\\
3.93939393939394	3.05044480809711e-61\\
3.98989898989899	3.14459929922022e-66\\
4.04040404040404	2.03661134593311e-71\\
4.09090909090909	8.28689254992142e-77\\
4.14141414141414	2.11843898192112e-82\\
4.19191919191919	3.40236509116158e-88\\
4.24242424242424	3.43310092487042e-94\\
4.29292929292929	2.17637100054421e-100\\
4.34343434343434	8.66802175452828e-107\\
4.39393939393939	2.16894109466786e-113\\
4.44444444444444	3.40970043234133e-120\\
4.49494949494949	3.36763794650903e-127\\
4.54545454545455	2.08965825488412e-134\\
4.5959595959596	8.14640208376858e-142\\
4.64646464646465	1.99524908054045e-149\\
4.6969696969697	3.07021719310871e-157\\
4.74747474747475	2.96812223256472e-165\\
4.7979797979798	1.80274860075821e-173\\
4.84848484848485	6.87906249332556e-182\\
4.8989898989899	1.64916475636579e-190\\
4.94949494949495	2.48393105888372e-199\\
5	2.35047552454136e-208\\
};
\addlegendentry{\textbf{Inf$+$1DR$+$RA}}

\end{axis}
\end{tikzpicture}%
    \end{minipage}
    \begin{minipage}{0.24\textwidth}
%
%
\definecolor{mycolor1}{rgb}{0.00000,0.44700,0.74100}%
\definecolor{mycolor2}{rgb}{0.85000,0.32500,0.09800}%
\begin{tikzpicture}
\tikzstyle{every node}=[font=\small,scale=0.75]

\begin{axis}[%
width=\linewidth,
height=0.4\linewidth,
scale only axis,
xmin=-1,
xmax=2,
ymin=0,
ymax=1.8,
xlabel={Position error \textit{($\log_{10}$  cm)}},
yticklabels={,,}
axis background/.style={fill=white},
axis lines=left,
legend style={legend cell align=left, align=left, draw=white!15!black}
]
\addplot [color=red,line width=1pt]
  table[row sep=crcr]{%
-1	3.25999666872112e-20\\
-0.96969696969697	3.58095285464417e-19\\
-0.939393939393939	3.66387439361341e-18\\
-0.909090909090909	3.49203163481258e-17\\
-0.878787878787879	3.1006474874649e-16\\
-0.848484848484849	2.56514039414692e-15\\
-0.818181818181818	1.97747162547049e-14\\
-0.787878787878788	1.42074064328687e-13\\
-0.757575757575758	9.51481575282336e-13\\
-0.727272727272727	5.94093331479547e-12\\
-0.696969696969697	3.45922706584335e-11\\
-0.666666666666667	1.87884244982158e-10\\
-0.636363636363636	9.5218765132481e-10\\
-0.606060606060606	4.50431250193621e-09\\
-0.575757575757576	1.98967286069473e-08\\
-0.545454545454545	8.21064482435617e-08\\
-0.515151515151515	3.16689180234134e-07\\
-0.484848484848485	1.14232708120525e-06\\
-0.454545454545455	3.85578097442989e-06\\
-0.424242424242424	1.21866171178706e-05\\
-0.393939393939394	3.60915581397896e-05\\
-0.363636363636364	0.00010022994369566\\
-0.333333333333333	0.000261208574844744\\
-0.303030303030303	0.000639309373653251\\
-0.272727272727273	0.00147064488120253\\
-0.242424242424242	0.00318214497411415\\
-0.212121212121212	0.00648180056391727\\
-0.181818181818182	0.012439347255333\\
-0.151515151515151	0.0225122713696763\\
-0.121212121212121	0.0384601975727215\\
-0.0909090909090909	0.0621040643869622\\
-0.0606060606060606	0.0949374615319173\\
-0.0303030303030303	0.137679878590611\\
0	0.189940901686718\\
0.0303030303030303	0.250186292168208\\
0.0606060606060606	0.316118532837158\\
0.0909090909090908	0.38541309515781\\
0.121212121212121	0.456559622971888\\
0.151515151515152	0.529450301035102\\
0.181818181818182	0.605410441035724\\
0.212121212121212	0.686569924241676\\
0.242424242424242	0.774739066467438\\
0.272727272727273	0.870163501436574\\
0.303030303030303	0.970609895943138\\
0.333333333333333	1.07115616575692\\
0.363636363636364	1.16484897946788\\
0.393939393939394	1.24409925590771\\
0.424242424242424	1.3023986190555\\
0.454545454545455	1.33577650720312\\
0.484848484848485	1.34348817332397\\
0.515151515151515	1.32774006577629\\
0.545454545454545	1.29268036470503\\
0.575757575757576	1.24317163141007\\
0.606060606060606	1.18384556993366\\
0.636363636363636	1.11864874400545\\
0.666666666666667	1.05075035582322\\
0.696969696969697	0.982544641504314\\
0.727272727272727	0.915604724055886\\
0.757575757575758	0.850669620601393\\
0.787878787878788	0.787837809893333\\
0.818181818181818	0.727006158350477\\
0.848484848484848	0.668357629781229\\
0.878787878787879	0.612589858637173\\
0.909090909090909	0.560711147625641\\
0.939393939393939	0.513523795013242\\
0.96969696969697	0.471144930015201\\
1	0.432905363772137\\
1.03030303030303	0.397719954786209\\
1.06060606060606	0.364710397821539\\
1.09090909090909	0.33369259997749\\
1.12121212121212	0.305219611957519\\
1.15151515151515	0.280139225690265\\
1.18181818181818	0.258914711223945\\
1.21212121212121	0.241098400328642\\
1.24242424242424	0.225265021992862\\
1.27272727272727	0.209457748242964\\
1.3030303030303	0.19192265271455\\
1.33333333333333	0.171764758173328\\
1.36363636363636	0.1492261323424\\
1.39393939393939	0.125509797267062\\
1.42424242424242	0.102306927300836\\
1.45454545454545	0.0812910023864874\\
1.48484848484848	0.063781333204314\\
1.51515151515152	0.0506219901751561\\
1.54545454545455	0.0421892316585745\\
1.57575757575758	0.0384089154086987\\
1.60606060606061	0.0387302514641559\\
1.63636363636364	0.0420976399855782\\
1.66666666666667	0.0470152110278824\\
1.6969696969697	0.0517734074544597\\
1.72727272727273	0.0548193629043806\\
1.75757575757576	0.0551561697972557\\
1.78787878787879	0.052610056724955\\
1.81818181818182	0.0478410996841596\\
1.84848484848485	0.0420773325199979\\
1.87878787878788	0.0366701656888024\\
1.90909090909091	0.0326410007079109\\
1.93939393939394	0.030383582718919\\
1.96969696969697	0.0296142348617924\\
2	0.0295612490124897\\
};
\addlegendentry{data1}

\addplot [color=blue,line width=1pt]
  table[row sep=crcr]{%
-1	5.29225674168648e-10\\
-0.96969696969697	3.90965738299155e-09\\
-0.939393939393939	2.58068643582646e-08\\
-0.909090909090909	1.52236520782626e-07\\
-0.878787878787879	8.02800674204075e-07\\
-0.848484848484849	3.78583584488402e-06\\
-0.818181818181818	1.5973476397214e-05\\
-0.787878787878788	6.03422758504943e-05\\
-0.757575757575758	0.000204289539885918\\
-0.727272727272727	0.000620662444077838\\
-0.696969696969697	0.00169537280234088\\
-0.666666666666667	0.00417461401820537\\
-0.636363636363636	0.0093002868604415\\
-0.606060606060606	0.0188401934280463\\
-0.575757575757576	0.0349386126387555\\
-0.545454545454545	0.0598303460326661\\
-0.515151515151515	0.0956081198893538\\
-0.484848484848485	0.144240032940009\\
-0.454545454545455	0.207807792933186\\
-0.424242424242424	0.288645882975771\\
-0.393939393939394	0.389029839803142\\
-0.363636363636364	0.510388352830785\\
-0.333333333333333	0.652391802069461\\
-0.303030303030303	0.81231779761797\\
-0.272727272727273	0.98482180612189\\
-0.242424242424242	1.16201454461344\\
-0.212121212121212	1.33379796165698\\
-0.181818181818182	1.48856525364833\\
-0.151515151515151	1.61435313134473\\
-0.121212121212121	1.70036410261012\\
-0.0909090909090909	1.73866700821968\\
-0.0606060606060606	1.72584950304778\\
-0.0303030303030303	1.66423780648204\\
0	1.56206622483577\\
0.0303030303030303	1.43212426231389\\
0.0606060606060606	1.28916589932843\\
0.0909090909090908	1.14714992596096\\
0.121212121212121	1.0173033737974\\
0.151515151515152	0.907074519621521\\
0.181818181818182	0.819389342975324\\
0.212121212121212	0.752102450677098\\
0.242424242424242	0.698481061831023\\
0.272727272727273	0.649520650042492\\
0.303030303030303	0.597533704638948\\
0.333333333333333	0.53914935580935\\
0.363636363636364	0.47606462441986\\
0.393939393939394	0.413432977601226\\
0.424242424242424	0.357198838805705\\
0.454545454545455	0.311908268399696\\
0.484848484848485	0.279739195464833\\
0.515151515151515	0.2605659599982\\
0.545454545454545	0.252420261423538\\
0.575757575757576	0.251845944676031\\
0.606060606060606	0.254148680815031\\
0.636363636363636	0.25397774968396\\
0.666666666666667	0.246582129209885\\
0.696969696969697	0.22941257212134\\
0.727272727272727	0.20310801926224\\
0.757575757575758	0.171060484408706\\
0.787878787878788	0.137720347122764\\
0.818181818181818	0.106727376787783\\
0.848484848484848	0.0799696324104609\\
0.878787878787879	0.0578267535038507\\
0.909090909090909	0.0399910996126283\\
0.939393939393939	0.026094857172035\\
0.96969696969697	0.0158394692386318\\
1	0.00883535274621803\\
1.03030303030303	0.00448720545986307\\
1.06060606060606	0.0020613329106067\\
1.09090909090909	0.000852719436820037\\
1.12121212121212	0.000316710027948634\\
1.15151515151515	0.00010540527595293\\
1.18181818181818	3.13936831304489e-05\\
1.21212121212121	8.36034665242044e-06\\
1.24242424242424	1.98954905212846e-06\\
1.27272727272727	4.22922964963679e-07\\
1.3030303030303	8.02835314919494e-08\\
1.33333333333333	1.36072139918969e-08\\
1.36363636363636	2.05889144430503e-09\\
1.39393939393939	2.78092973557605e-10\\
1.42424242424242	3.36079043322706e-11\\
1.45454545454545	4.5202488401272e-12\\
1.48484848484848	9.41673505951117e-12\\
1.51515151515152	8.04792658044485e-11\\
1.54545454545455	6.36823273107441e-10\\
1.57575757575758	4.49884011042755e-09\\
1.60606060606061	2.83643742385438e-08\\
1.63636363636364	1.59600911151485e-07\\
1.66666666666667	8.01469542134234e-07\\
1.6969696969697	3.59193257370529e-06\\
1.72727272727273	1.43667596166695e-05\\
1.75757575757576	5.12836551818677e-05\\
1.78787878787879	0.000163376133697881\\
1.81818181818182	0.000464502176915868\\
1.84848484848485	0.00117862701094409\\
1.87878787878788	0.00266903655801074\\
1.90909090909091	0.00539413922619141\\
1.93939393939394	0.0097292480855889\\
1.96969696969697	0.0156612311715317\\
2	0.0224989380223443\\
};
\addlegendentry{data2}
\legend{}
\end{axis}
\end{tikzpicture}%
    \end{minipage}
    \caption{\textit{Left:} Qualitative example of rig initializations before final refinement. \textit{Right:} Distribution of rotation and translation errors before final refinement.
    }
    \label{fig:initial_eval}
\end{figure}

In this section we evaluate the effect of delaying the estimation of the intrinsic parameters on the quality of the initial estimates, i.e.~before running bundle adjustment. 
Similar to the evaluation for robustness in Section~\ref{sec:robust}, we run the different methods on multiple sub-sequences and evaluate the extrinsics error of the initial estimates. 
A qualitative example of the extrinsics is shown in Figure~\ref{fig:initial_eval} (Left) and it is obvious that the extrinsics estimate for \textbf{Inf$+$1DR$+$RA} is much better and almost close to the final result. 
Figure~\ref{fig:initial_eval} (Right) shows the distribution of the extrinsics error for both methods, where \textbf{Inf$+$1DR$+$RA} outperforms \textbf{Inf$+$RD$+$RA} especially in position error. However, as shown in Table~\ref{tbl:robustness} most of these initial errors can be recovered in the final refinement.

\subsection{Evaluation on RobotCar Dataset}

\begin{figure}[tbp]
\centering
\subfigure[]{
\label{fig:robotcar-extrinsic-diff}
\includegraphics[trim=0 0 0 0,clip,width=0.30\textwidth]{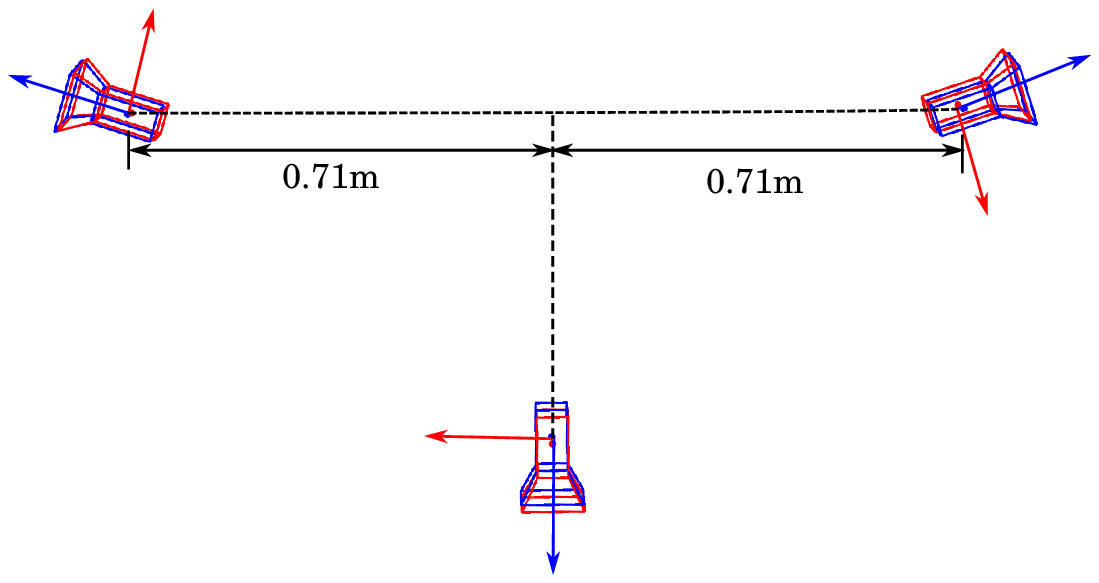}}
\subfigure[]{
\label{fig:robotcar-raw-image}
\includegraphics[width=0.18\textwidth]{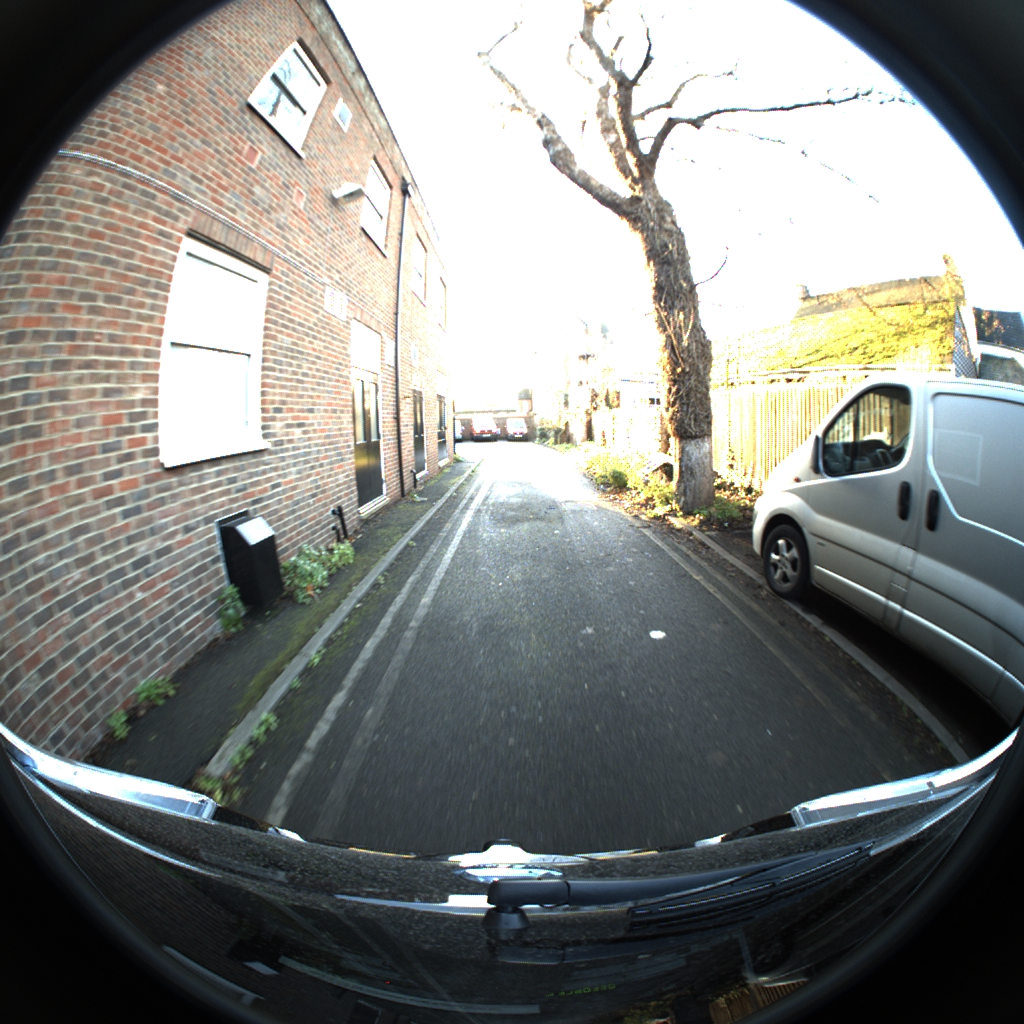}}
\subfigure[]{
\label{fig:robotcar-undistort}
\includegraphics[trim=60 60 60 60,clip,width=0.18\textwidth]{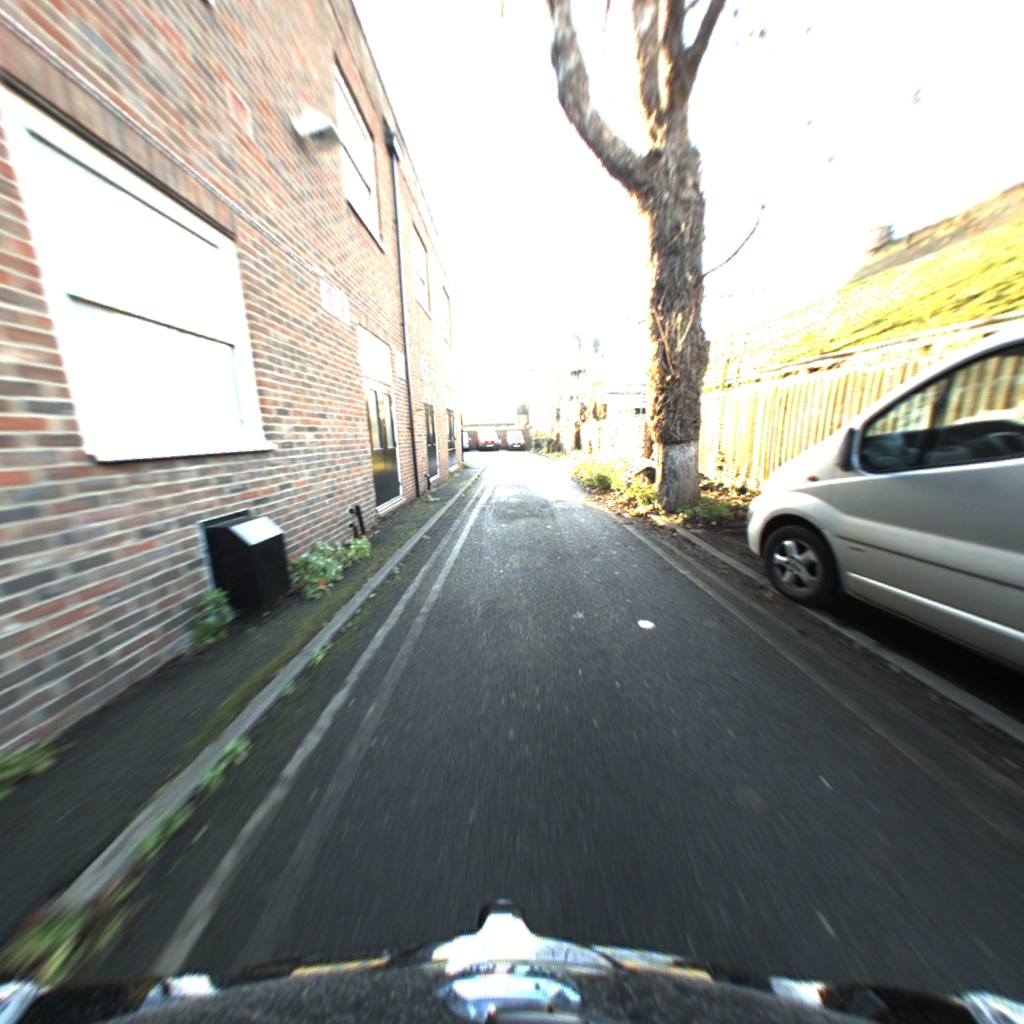}}
\subfigure[]{
\label{fig:robotcar-undistort-gt}
\includegraphics[width=0.18\textwidth]{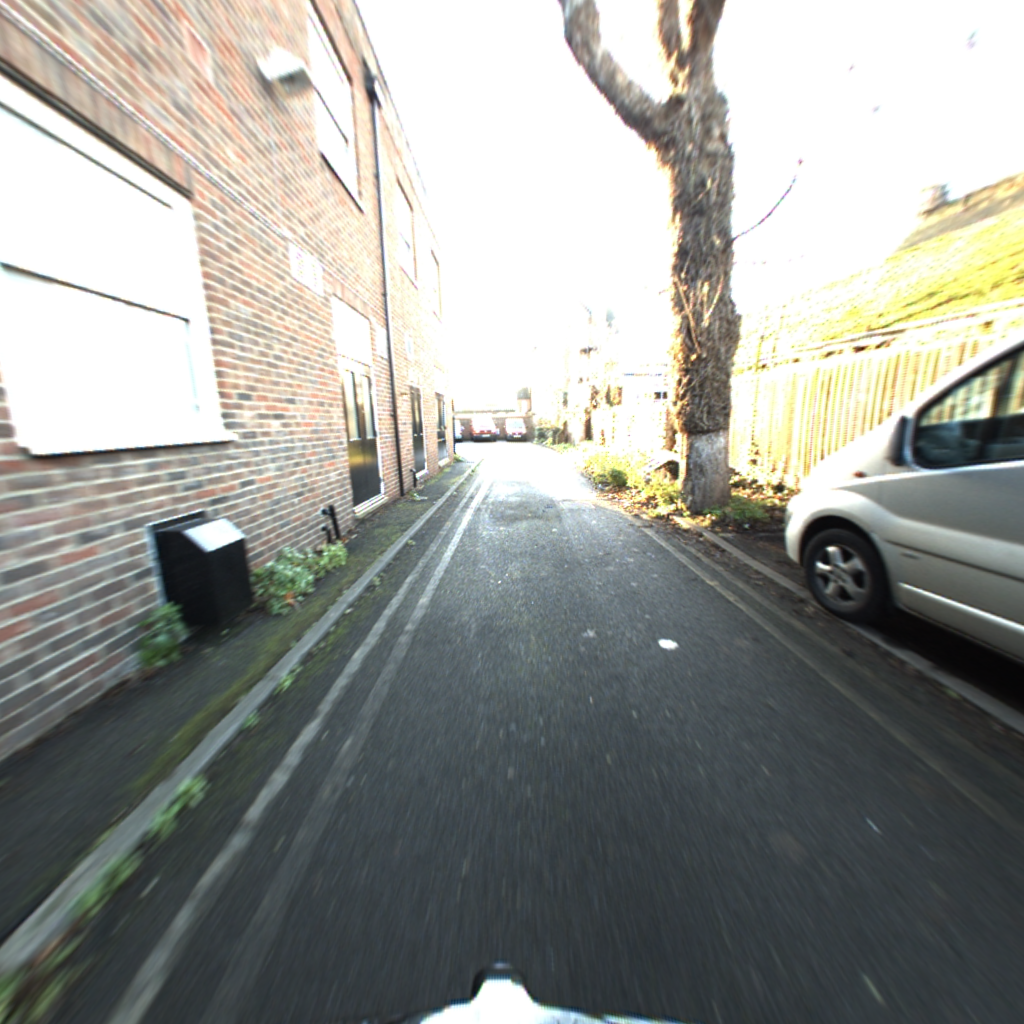}}
\caption{\textbf{Results on RobotCar datasets.} The extrinsics for out method(blue) and groundtruth(red) are plotted in (a). To validate the intrinsics, the raw image (b) is undistorted using our calibrated results (c) and groundtruth parameters (d).}
\label{fig:robotcar}
\end{figure}

In addition to the experiments mentioned above, we evaluate our calibration method on the public  benchmark RobotCar Dataset~\cite{RCDRTKArXiv}.
We select a short sequence of 30 seconds from the 2014/12/16 datasets (frame No.500 to frame No.900) recorded in the morning to calibrate the three Grasshopper2 cameras pointing left, back and right respectively. The map and calibration groundtruth is obtained from the RobotCar Seasons Dataset~\cite{sattler2018benchmarking}.
The calibration takes only 3 minutes on a normal PC and the extrinsic results are shown in Figure~\ref{fig:robotcar-extrinsic-diff}.
The position error is 1.04cm and rotation error is 0.213$^\circ$. 
To validate the intrinsic parameters, 
we compare the results directly from undistorting the raw image  Figure~\ref{fig:robotcar-raw-image}. Figure~\ref{fig:robotcar-undistort} and Figure~\ref{fig:robotcar-undistort-gt} are the undistorted image for our method and the groundtruth respectively.
Although this benchmark is designed for visual localization and place recognition algorithms under changing conditions, we show our method robustly and accurately estimates the camera calibration parameters even with real vehicle vision data in urban roads.

\subsection{Application: Robot Localization in a Garden}
\begin{figure}[tb]
\centering
\subfigure[Extrinsics]{
\label{fig:extrinsic-diff}
\includegraphics[trim=60 10 130 10,clip,width=0.30\textwidth]{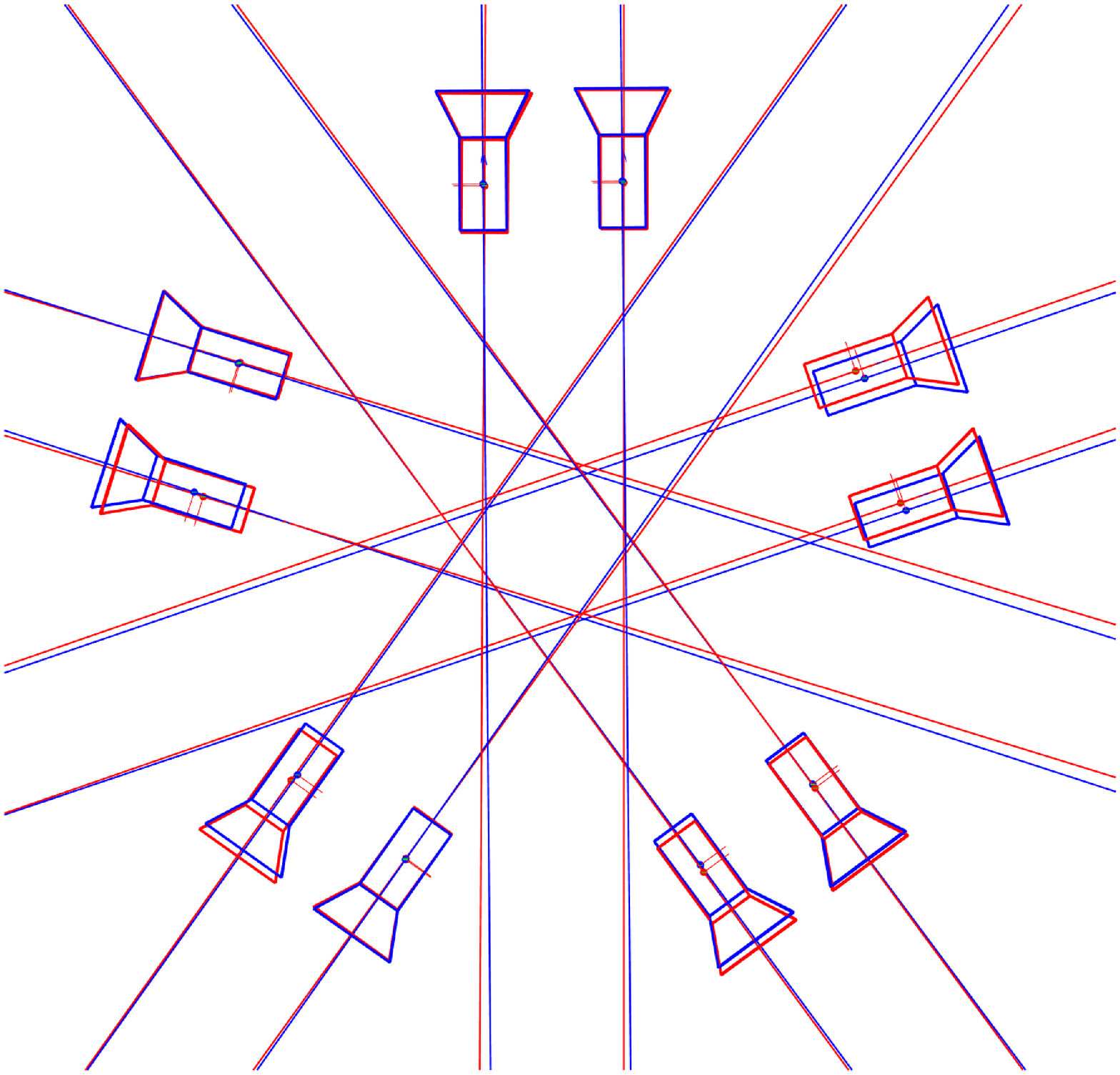}}
\subfigure[Easy case]{
\label{fig:slam-result-easy}
\includegraphics[width=0.30\textwidth]{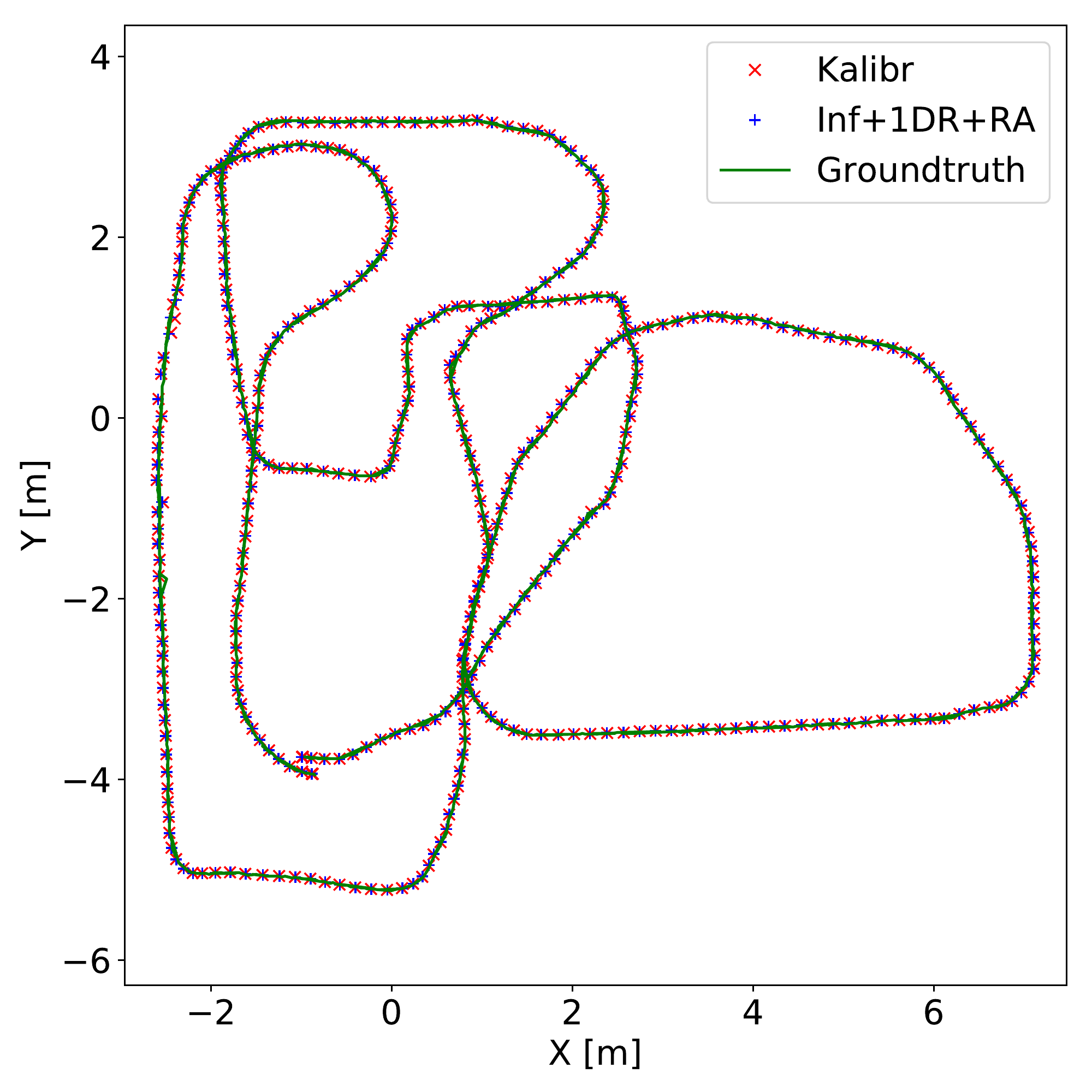}}
\subfigure[Hard case]{
\label{fig:slam-result-hard}
\includegraphics[width=0.30\textwidth]{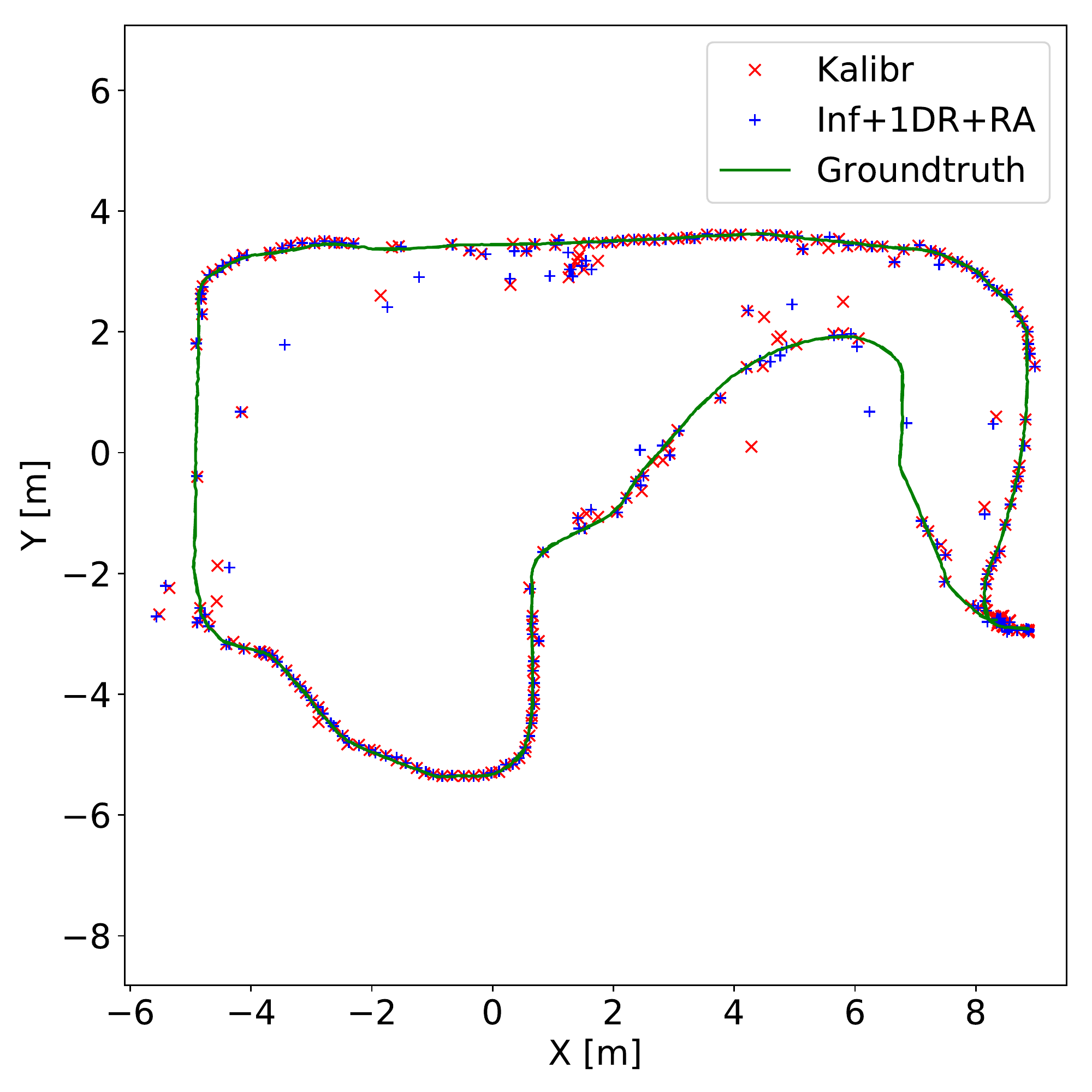}}
\caption{\textbf{Results in gardening datasets.} The extrinsics are plotted in (a). (b) shows the localization trajectory of an easy dataset and (c) a hard one. The Kalibr results are indicated by red and our method by blue.}
\label{fig:garden}
\end{figure}

Finally we evaluate our proposed framework in a real robotics application, namely localization in an outdoor environment. We attach the pentagonal rig to a small robot which autonomously navigates in a garden. We record several datasets of the robot driving around in the garden. From one of the recordings we build a map using the calibration obtained from Aprilgrid calibration with Kalibr. We then calibrate the camera rig using one of the other datasets and evaluate localization performance on the rest of the datasets. 
The position of the robot is tracked with a TopCon laser tracker yielding accurate position used as groundtruth. 
The plot of Kalibr extrinsics and results from our results shown in Figure~\ref{fig:extrinsic-diff} confirms the high accuracy of our calibration method. 
In Figure~\ref{fig:slam-result-easy} and Figure~\ref{fig:slam-result-hard} we plot the localized trajectory of two different localization datasets using the calibration results of Kalibr and our method. The median position errors for the two sequences are 3.56 cm and 9.22 cm using results from the proposed method, and 3.77 cm and 9.67 cm using calibration with Kalibr.
Using a calibration estimated from the  map we are able to achieve slightly higher accuracy for localization compared to the pattern-based approach. 

\section{Conclusions}
We have proposed a method for complete calibration, both intrinsic and extrinsic, of multi-camera systems.
Due to the use of natural scene features, our calibration method can be used in any arbitrary indoor and outdoor environments without the aid of other calibration patterns or setups. 
The extensive experiments and real case application demonstrate the high accuracy, efficiency and robustness of our proposed calibration method.
Given the practical usefulness of our approach, we expect it to have large impact in the robotics and autonomous vehicle community. 

\PAR{Acknowledgement}  This work was supported by the Swedish Foundation for Strategic Research (Semantic Mapping and Visual Navigation for Smart Robots), the Chalmers AI Research Centre (CHAIR) (VisLocLearn), OP VVV project Research Center for Informatics No.\ CZ.02.1.01/0.0/0.0/$16\_019$/0000765, and EU Horizon 2020 research and innovation program under grant No. 688007 (TrimBot2020).
Viktor Larsson was supported by an ETH Z\"urich Postdoctoral Fellowship. 
\bibliographystyle{splncs04}
\bibliography{main}

\begin{thebibliography}{10}
\providecommand{\url}[1]{\texttt{#1}}
\providecommand{\urlprefix}{URL }
\providecommand{\doi}[1]{https://doi.org/#1}

\bibitem{agarwal2018ceres}
Agarwal, S., Mierle, K., et~al.: Ceres solver, 2013. URL http://ceres-solver.
  org  (2018)

\bibitem{arth2009wide}
Arth, C., Wagner, D., Klopschitz, M., Irschara, A., Schmalstieg, D.: Wide area
  localization on mobile phones. In: 2009 8th ieee international symposium on
  mixed and augmented reality. pp. 73--82. IEEE (2009)

\bibitem{bujnak2008general}
Bujnak, M., Kukelova, Z., Pajdla, T.: A general solution to the p4p problem for
  camera with unknown focal length. In: Computer Vision and Pattern Recognition
  (CVPR) (2008)

\bibitem{camposeco2015non}
Camposeco, F., Sattler, T., Pollefeys, M.: Non-parametric structure-based
  calibration of radially symmetric cameras. In: International Conference on
  Computer Vision (ICCV) (2015)

\bibitem{carrera2011slam}
Carrera, G., Angeli, A., Davison, A.J.: Slam-based automatic extrinsic
  calibration of a multi-camera rig. In: International Conference on Robotics
  and Automation (ICRA) (2011)

\bibitem{fischler1981random}
Fischler, M.A., Bolles, R.C.: Random sample consensus: a paradigm for model
  fitting with applications to image analysis and automated cartography.
  Communications of the ACM  \textbf{24}(6),  381--395 (1981)

\bibitem{geppert2019efficient}
Geppert, M., Liu, P., Cui, Z., Pollefeys, M., Sattler, T.: Efficient 2d-3d
  matching for multi-camera visual localization. In: International Conference
  on Robotics and Automation (ICRA) (2019)

\bibitem{govindu2006robustness}
Govindu, V.M.: Robustness in motion averaging. In: Asian Conference on Computer
  Vision (ACCV) (2006)

\bibitem{haralick1994review}
Haralick, B.M., Lee, C.N., Ottenberg, K., N{\"o}lle, M.: Review and analysis of
  solutions of the three point perspective pose estimation problem.
  International journal of computer vision  \textbf{13}(3),  331--356 (1994)

\bibitem{hartley2003multiple}
Hartley, R., Zisserman, A.: Multiple view geometry in computer vision.
  Cambridge university press (2003)

\bibitem{heng2019project}
Heng, L., Choi, B., Cui, Z., Geppert, M., Hu, S., Kuan, B., Liu, P., Nguyen,
  R., Yeo, Y.C., Geiger, A., et~al.: Project autovision: Localization and 3d
  scene perception for an autonomous vehicle with a multi-camera system. In:
  International Conference on Robotics and Automation (ICRA) (2019)

\bibitem{heng2015leveraging}
Heng, L., Furgale, P., Pollefeys, M.: Leveraging image-based localization for
  infrastructure-based calibration of a multi-camera rig. Journal of Field
  Robotics  \textbf{32}(5),  775--802 (2015)

\bibitem{heng2015self}
Heng, L., Lee, G.H., Pollefeys, M.: Self-calibration and visual slam with a
  multi-camera system on a micro aerial vehicle. Autonomous robots
  \textbf{39}(3),  259--277 (2015)

\bibitem{heng2013camodocal}
Heng, L., Li, B., Pollefeys, M.: Camodocal: Automatic intrinsic and extrinsic
  calibration of a rig with multiple generic cameras and odometry. In:
  International Conference on Intelligent Robots and Systems (IROS) (2013)

\bibitem{josephson2009pose}
Josephson, K., Byrod, M.: Pose estimation with radial distortion and unknown
  focal length. In: Computer Vision and Pattern Recognition (CVPR) (2009)

\bibitem{kannala2006generic}
Kannala, J., Brandt, S.S.: A generic camera model and calibration method for
  conventional, wide-angle, and fish-eye lenses. Trans. Pattern Analysis and
  Machine Intelligence (PAMI)  \textbf{28}(8),  1335--1340 (2006)

\bibitem{kukelova2013real}
Kukelova, Z., Bujnak, M., Pajdla, T.: Real-time solution to the absolute pose
  problem with unknown radial distortion and focal length. In: International
  Conference on Computer Vision (ICCV) (2013)

\bibitem{kukelova2016efficient}
Kukelova, Z., Heller, J., Fitzgibbon, A.: Efficient intersection of three
  quadrics and applications in computer vision. In: Computer Vision and Pattern
  Recognition (CVPR) (2016)

\bibitem{kumar2008simple}
Kumar, R.K., Ilie, A., Frahm, J.M., Pollefeys, M.: Simple calibration of
  non-overlapping cameras with a mirror. In: Computer Vision and Pattern
  Recognition (CVPR) (2008)

\bibitem{larsson2017making}
Larsson, V., Kukelova, Z., Zheng, Y.: Making minimal solvers for absolute pose
  estimation compact and robust. In: International Conference on Computer
  Vision (ICCV) (2017)

\bibitem{larsson2018camera}
Larsson, V., Kukelova, Z., Zheng, Y.: Camera pose estimation with unknown
  principal point. In: Computer Vision and Pattern Recognition (CVPR) (2018)

\bibitem{larsson2019revisiting}
Larsson, V., Sattler, T., Kukelova, Z., Pollefeys, M.: Revisiting radial
  distortion absolute pose. In: International Conference on Computer Vision
  (ICCV) (2019)

\bibitem{li2013multiple}
Li, B., Heng, L., Koser, K., Pollefeys, M.: A multiple-camera system
  calibration toolbox using a feature descriptor-based calibration pattern. In:
  International Conference on Intelligent Robots and Systems (IROS) (2013)

\bibitem{liu2018towards}
Liu, P., Geppert, M., Heng, L., Sattler, T., Geiger, A., Pollefeys, M.: Towards
  robust visual odometry with a multi-camera system. In: International
  Conference on Intelligent Robots and Systems (IROS) (2018)

\bibitem{lowe2004distinctive}
Lowe, D.G.: Distinctive image features from scale-invariant keypoints.
  International Journal of Computer Vision (IJCV)  \textbf{60}(2),  91--110
  (2004)

\bibitem{RCDRTKArXiv}
Maddern, W., Pascoe, G., Gadd, M., Barnes, D., Yeomans, B., Newman, P.:
  Real-time kinematic ground truth for the oxford robotcar dataset. arXiv
  preprint arXiv: 2002.10152  (2020), \url{https://arxiv.org/pdf/2002.10152}

\bibitem{maye2013self}
Maye, J., Furgale, P., Siegwart, R.: Self-supervised calibration for robotic
  systems. In: 2013 IEEE Intelligent Vehicles Symposium (IV). pp. 473--480.
  IEEE (2013)

\bibitem{olson2011apriltag}
Olson, E.: Apriltag: A robust and flexible visual fiducial system. In:
  International Conference on Robotics and Automation (ICRA) (2011)

\bibitem{olsson2011stable}
Olsson, C., Enqvist, O.: Stable structure from motion for unordered image
  collections. In: Scandinavian Conference on Image Analysis (SCIA) (2011)

\bibitem{penate2013exhaustive}
Penate-Sanchez, A., Andrade-Cetto, J., Moreno-Noguer, F.: Exhaustive
  linearization for robust camera pose and focal length estimation. Trans.
  Pattern Analysis and Machine Intelligence (PAMI)  \textbf{35}(10),
  2387--2400 (2013)

\bibitem{robinson2017robust}
Robinson, A., Persson, M., Felsberg, M.: Robust accurate extrinsic calibration
  of static non-overlapping cameras. In: International Conference on Computer
  Analysis of Images and Patterns (CAIP) (2017)

\bibitem{sattler2018benchmarking}
Sattler, T., Maddern, W., Toft, C., Torii, A., Hammarstrand, L., Stenborg, E.,
  Safari, D., Okutomi, M., Pollefeys, M., Sivic, J., et~al.: Benchmarking 6dof
  outdoor visual localization in changing conditions. In: Computer Vision and
  Pattern Recognition (CVPR) (2018)

\bibitem{schonberger2016structure}
Schonberger, J.L., Frahm, J.M.: Structure-from-motion revisited. In: Computer
  Vision and Pattern Recognition (CVPR) (2016)

\bibitem{schwesinger2016automated}
Schwesinger, U., B{\"u}rki, M., Timpner, J., Rottmann, S., Wolf, L., Paz, L.M.,
  Grimmett, H., Posner, I., Newman, P., H{\"a}ne, C., et~al.: Automated valet
  parking and charging for e-mobility. In: Intelligent Vehicles Symposium (IV).
  IEEE (2016)

\bibitem{sorkine2017least}
Sorkine-Hornung, O., Rabinovich, M.: Least-squares rigid motion using svd.
  Computing  \textbf{1}(1) (2017)

\bibitem{strisciuglio2018trimbot2020}
Strisciuglio, N., Tylecek, R., Blaich, M., Petkov, N., Biber, P., Hemming, J.,
  van Henten, E., Sattler, T., Pollefeys, M., Gevers, T., et~al.: Trimbot2020:
  an outdoor robot for automatic gardening. In: ISR 2018; 50th International
  Symposium on Robotics. pp.~1--6. VDE (2018)

\bibitem{sturm1999plane}
Sturm, P.F., Maybank, S.J.: On plane-based camera calibration: A general
  algorithm, singularities, applications. In: Computer Vision and Pattern
  Recognition (CVPR) (1999)

\bibitem{thirthala2012radial}
Thirthala, S., Pollefeys, M.: Radial multi-focal tensors. International Journal
  of Computer Vision (IJCV)  \textbf{96}(2),  195--211 (2012)

\bibitem{triggs1999camera}
Triggs, B.: Camera pose and calibration from 4 or 5 known 3d points. In:
  International Conference on Computer Vision (ICCV) (1999)

\bibitem{tsai1987versatile}
Tsai, R.: A versatile camera calibration technique for high-accuracy 3d machine
  vision metrology using off-the-shelf tv cameras and lenses. IEEE Journal on
  Robotics and Automation  \textbf{3}(4),  323--344 (1987)

\bibitem{wu2015p3}
Wu, C.: P3. 5p: Pose estimation with unknown focal length. In: Computer Vision
  and Pattern Recognition (CVPR) (2015)

\bibitem{zhang2004extrinsic}
Zhang, Q., Pless, R.: Extrinsic calibration of a camera and laser range finder
  (improves camera calibration). In: International Conference on Intelligent
  Robots and Systems (IROS) (2004)

\bibitem{zheng2014general}
Zheng, Y., Sugimoto, S., Sato, I., Okutomi, M.: A general and simple method for
  camera pose and focal length determination. In: Computer Vision and Pattern
  Recognition (CVPR) (2014)

\end{thebibliography}
\end{document}